\documentclass{article}
\usepackage{amssymb}
\usepackage{booktabs}  
\usepackage{caption}   
\usepackage{graphicx}  
\usepackage{amsmath}  
\usepackage{array}
\usepackage{multirow}

\PassOptionsToPackage{verbose=true,letterpaper,margin=1in}{geometry}
\usepackage{geometry}

\usepackage{listings}

\usepackage[table]{xcolor}

\definecolor{codebg}{rgb}{0.98,0.98,0.98}       
\definecolor{vscComment}{rgb}{0.0, 0.5, 0.0}     
\definecolor{vscKeyword}{rgb}{0.0, 0.0, 0.8}     
\definecolor{vscString}{rgb}{0.65, 0.13, 0.13}   
\definecolor{vscBlack}{rgb}{0.1, 0.1, 0.1}       

\lstdefinestyle{vscode}{
  backgroundcolor=\color{codebg},
  basicstyle=\ttfamily\small\color{vscBlack},
  commentstyle=\color{vscComment}\itshape,
  keywordstyle=\color{vscKeyword}\bfseries,
  stringstyle=\color{vscString},
  breaklines=true,
  frame=single,
  captionpos=b,
  language=Python,
  morekeywords={Cnfg},
}

\usepackage[preprint]{corl_2025} 

\title{McARL:Morphology-Control-Aware Reinforcement Learning for Generalizable Quadrupedal Locomotion}

\author{
  Prakhar Mishra\textsuperscript{1}\thanks{\textcolor{magenta}{Videos and code:} \url{https://prakharmishra27.github.io/McARL/}},
  Amir Hossain Raj\textsuperscript{2},
  Xuesu Xiao\textsuperscript{2},
  Dinesh Manocha\textsuperscript{1} \\
  \textsuperscript{1}University of Maryland, College Park \\
  \textsuperscript{2}George Mason University
}

\begin{document}
\maketitle


\begin{abstract}
 We present Morphology-Control-Aware Reinforcement Learning (McARL), a new approach to overcome challenges of hyperparameter tuning and transfer loss, that enables generalizable locomotion across robot morphologies. We  use a morphology conditioned policy by incorporating a randomized morphological vector, sampled from the defined morphology range, into both the actor and critic. This helps the policy to learn the parameters that can be helpful to control the robots that have similar characteristics. 
    We demonstrate that a single policy trained on a Unitree Go1 robot using McARL can be transferred to a different morphology (e.g., Unitree Go2 robot) and can achieve zero-shot transfer velocity of as high as 3.5 m/s without retraining or fine-tuning.  Moreover, it achieves 6.0 m/s on the training Go1 robot and generalizes to various other morphologies like A1 and Mini-Cheetah robots. We also analyze the impact on the transfer rate as a function of the morphology distance and highlight the benefits over prior approaches. McARL achieves 44–150\% higher transfer performance on Go2, Mini Cheetah and A1 compared to other variants of PPO, as discussed in results sections.

\end{abstract}



\section{Introduction}

    Recent advances in reinforcement learning methods have significantly improved the performance of legged-locomotion controllers \cite{margolis2024rapid}, \cite{kumar2021rma}, \cite{rudin2022learning}, \cite{lee2020learning}. Reinforcement learning provides robust, scalable and adaptive controllers, that's why they have gained traction compared to traditional controllers \cite{bledt2018cheetah}, \cite{bosworth2016robot}, \cite{ding2019real}, \cite{herzog2016structured}. While in these traditional model-based controller, the success is highly dependent on the design and invention of the reduced order models, in contrast Reinforcement learning doesn't require such time-intensive and iterative methodologies. 
    However, most of these RL controllers are trained for specific robot morphology and tasks, and may not generalize well across multiple morphologies \cite{margolis2024rapid}, \cite{kumar2021rma}.  In order to implement the same controller on a different robot, they needs to be trained again using the same policy network. This process may involve fine-tuning the hyper-parameters, which can be time consuming.
    There has been some progress to address such challenges like system identification \cite{yu2019sim} and domain randomization \cite{tan2018sim} etc, but such techniques still do not generalize to unseen morphologies and the key limitations of such methods is treating morphology features as noise, instead of some learnable features.
    
    Given the similar morphological characteristics of legged robots, which include four limbs, similar leg lengths, leg mass, DOF, etc., it may be useful to incorporate these characteristics into the learning method to overcome the challenges of zero-shot transfer, hyper-parameter tuning and generalization. A generalizable approach can help the users to use a pretrained models and deploy it on different robotic platform. Recent works have successfully tried to integrate the morphology in the RL for learning locomotion controller, but with various limitation and lacking robustness \cite{feng2023genloco}, \cite{sferrazza2024body}, \cite{shafiee2024manyquadrupeds},\cite{bohlinger2024one}.


    \noindent {\bf Main Results:} We present a novel approach for generalizable quadrupedal locomotion (McARL). Our approach models the controller for a range of morphologies by incorporating a learned morphology embedding ($z_{m}$) in the actor-critic policy, which learn robot specific charateristics like leg length, leg mass, base mass and joint torque limits etc. detailed in \ref{tab:morph-params}. While training each environment is assigned a random vector, given by $m \sim \rho(m)$, and then these vectors are passed through an encoder to lean $z_{m}$, which is used as an input to the policy. These randomized vectors, $m$ are the 14-D representation of the robot control and morphology parameters.
    The novel components of our work include:

\begin{itemize}
    \item \textbf{Morph condition policy }: We present a novel Morphology condition policy learning for legged locomotion for zero-shot transfer and minimal-hyperparameter tuning.
     \item \textbf{Zero-shot Transfer}: We demonstrate good sim-to-sim transfer behavior for McARL, when we train on just one go1 robot (which achieved 6m/sec) and zero-shot transfer to go2 (which achieved 3.5m/s), Mini Cheetah (1.5m/s) and also A1 robot that maintains a standing posture without failing.
    \item \textbf{History-Aware Curriculum}: We integrate history-aware in curriculum learning and highlight the benefits for morphology-control-aware reinforcement learning.
     \item \textbf{Empirical Evaluation}: We perform a comprehensive analysis of McARL with state of the art methods and compare the performance with  reinforcement learning methods. We demonstrate the effectiveness of incorporating morphology input in the policy learning as McARL achieves 44–150\% higher transfer performance.
     \item \textbf{Real-world Demonstration}: We have implemented our policy on a go1 robot, trained on separate morphology, with minimal finetuning. We observed stability, after some finetuning, and better responsiveness than the baselines, most importantly to a completely separate morphology on which its not even trained, an important capability not demonstrated by prior work.
  
\end{itemize}

\begin{figure}[ht]
    \centering
    \includegraphics[width=0.75\linewidth]{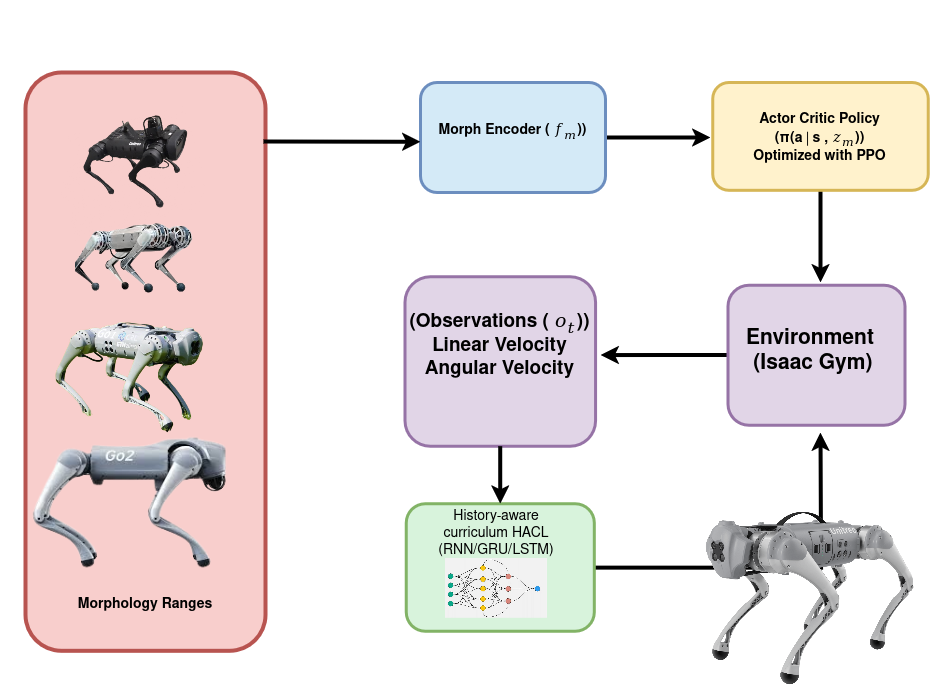}
    \caption{Overview of our Morphology-Control-Aware Reinforcement Learning (McARL), a novel framework incorporates 14-dimensional morphology vector into the PPO policy. }
    \label{fig:example-image}
\end{figure}


\section{Related Work}
\label{sec:related work}


\textbf{Model-based control:} Simplified models and hand-specified gaits have been used for legged locomotion to achieve stable gait and balance \cite{bledt2018cheetah}, \cite{bosworth2016robot}, \cite{ding2019real}, \cite{herzog2016structured}. Subsequent layered expansion has been done on these model-based controller, that are capable of traversing soft, rough and slippery terrains etc \cite{fahmi2020stance}, \cite{kuindersma2016optimization}. But recent developments address specific limitations like whole body control, regulating ground reaction forces or predictive control \cite{bledt2018cheetah}, \cite{dai2014whole}, \cite{kim2019highly}. McARL achieves robust and stable behavior by incorporating control parameter in the $z_m$ latent, but it can lead to conservative results.


\textbf{Sim-to-Real and Zero-shot Transfer:} A lot of work has been done in this direction to achieve complex locomotion behavior in real world using techniques like domain randomization \cite{tobin2017domain}, where the policy is trained over a wide variety of parameter and sensor noises to achieve a robust real world behavior \cite{tan2018sim}, \cite{xie2021dynamics}, \cite{peng2018sim}, \cite{nachum2019multi}. However, domain randomization leads to conservative policy behaviors, probably due to treating it like noise instead of as learnable parameters\cite{kumar2021rma}, \cite{luo2017robust}. Other ways to reduce the Sim-to-Real gap is to either make the simulation as accurate as possible to the real world or filling the gap by piece by piece motor data from the real world \cite{hwangbo2019learning}, \cite{tan2018sim}. There has been some progress made in legged robots in terms of zero-shot transfer, but achieving robustness still remains challenging \cite{zhao2024zsl}, \cite{li2022zero}, \cite{pan2020zero}. Our approach McARL instead of treating morphology as noise, uses as a learnable parameter and can be deployed on real world with minimal tuning.



\textbf{Morphology-conditioned Learning:} Recent development in legged locomotion have attracted a lot of attention towards morphology. Ranging from employing transformers  \cite{sferrazza2024body}  to graph neural networks \cite{rafiei2022ieee}, \cite{scarselli2008graph} to history-based inference of sensory input data \cite{feng2023genloco}. While transformers \cite{vaswani2017attention} were applied for various natural language application but later diversified to other areas like , audio processing, computer vision \cite{dosovitskiy2020image} and later to legged robots \cite{sferrazza2024body}. While there has a been some progress made in incorporating the morphology as a learnable feature and optimize the locomotion learning skills \cite{feng2023genloco}, \cite{huang2020one}, \cite{chiappa2022dmap}, \cite{gupta2022metamorph}, but none of these approaches explicitly condition policy learning on the morphology, a gap, which McARL fills.

\textbf{Curriculum Learning in RL:}
\citet{bengio2009curriculum} propose rule-based update method similar to continuation methods where training happens in a progression, starting with easier examples, then further increasing the difficulty of the training. As per \cite{wang2021survey}, there are following types of curriculum namely, Self-paced Learning, Transfer Teacher, RL Teacher and other automatic curriculum learning \cite{matiisen2019teacher}. Various curriculum techniques have been employed for legged lcomotion ranging from multi-terrain curriculum approach, game-inspired and fixed rule based or box-adaptive curriculum for learning locomotion skills \cite{aractingi2023controlling}, \cite{margolis2024rapid}, \cite{rudin2022learning}. But most of these methods employed a fixed rule based curriculum update method and ignore the importance of history in legged locomotion, McARL utilizes the history-aspect of the locomotion \cite{mishrahacl}




\section{McARL: Morphology-Control-Aware Reinforcement Learning}
\label{sec:citations}
The main goals of our approach are: (i) to train a single policy which can be trained using any single robot; (ii) to deploy or have the capacity for zero-shot transfer to unseen morphologies; (iii) not use any special or privileged observation during training or deployment. The robot morphology can be represented as a parameterized MDP \eqref{eq:morph-mdp}, where $P_{m}$ acts as the transition model and  $r_{m}$ as the reward function for the particular morphology of the robot, where $m \in \mathcal{R} \subset \mathbb{R}^{d_m}$

\begin{equation}
\mathcal{M}(m) = \langle \mathcal{S}, \mathcal{A}, P_m, r_m, \gamma \rangle,\quad m \in \mathcal{R} \subset \mathbb{R}^{d_m}.
\label{eq:morph-mdp}
\end{equation}
We develop   McARL as a framework to solve this meta-MDP, where the policy must learn to solve it. In the rest of the section, we provide more details about the morphology randomization, encoder, policy architecture, the PPO optimization and the History-Aware Curriculum Learning (HACL).



\begin{table}[ht]
\centering
\caption{Morphological and control parameters used for vector generation ($z_\text{morph}$)}
\label{tab:morph-params}
\resizebox{\linewidth}{!}{%
\begin{tabular}{|>{\raggedright}p{3.5cm}|>{\raggedright}p{4cm}|
                >{\raggedright}p{3.5cm}|>{\raggedright\arraybackslash}p{4cm}|}
\hline
\textbf{Morphology Parameters} & \textbf{Min \& Max Values} & \textbf{Morphology Parameters} & \textbf{Min \& Max Values} \\
\hline
Hip-to-thigh length & $[0.0,\ 0.08]$ m & Thigh-to-calf length & $[0.2,\ 0.213]$ m \\
\hline
Calf-to-foot length & $[0.0,\ 0.213]$ m & Thigh mass & $[0.634,\ 1.152]$ kg \\
\hline
Calf mass & $[0.064,\ 0.166]$ kg & Hip mass & $[0.510,\ 0.696]$ kg \\
\hline
Foot mass & $[0.0,\ 0.06]$ kg & Base mass & $[3.3,\ 6.921]$ kg \\
\hline
Hip joint  & $[-1.6,\ -0.803]$ rad & Hip joint  & $[0.803,\ 1.6]$ rad \\
\hline
Thigh joint & $[-2.6,\ -1.047]$ rad & Thigh joint  & $[2.6,\ 4.189]$ rad \\
\hline
Calf joint & $[-2.723,\ -2.6]$ rad & Calf joint  & $[-0.916,\ 2.6]$ rad \\
\hline
\textbf{Control Parameters} & \textbf{Min \& Max Values} & \textbf{Control Parameters} & \textbf{Min \& Max Values} \\
\hline
Joint stiffness & $[20.0,\ 30.0]$ Nm/rad & Joint damping & $[0.35,\ 0.55]$ Nms/rad \\
\hline
Action scale & $[0.20,\ 0.45]$ & Hip scale  & $[0.3,\ 0.5]$ \\
\hline
Motor strength  & $[0.9,\ 1.1]$ & Hip torque  & $[16,\ 34]$ Nm \\
\hline
Thigh torque  & $[16,\ 34]$ Nm & Calf torque  & $[24,\ 46]$ Nm \\
\hline
\end{tabular}
}
\end{table}

\subsection{Morphology Randomization and Encoder}


\textbf{Morphology:} Our goal to develop a morphology-aware controller is select the right family of legged robots and based on those robots, we should define the morphology and control parameters, which will be used to generate the morphology-control vector. In our case we consider these robots: Unitree Go1, Go2, Mini-cheetah and A1. The ranges for morphology and control parameters as detailed in Table 1. Based on Table 1 parameters ranges, we generate a separate random vector of dimension 14 for all 4000 training environments in Isaac gym and is given by: $m \sim \rho(m) = \text{Uniform}[m_{\text{min}}, m_{\text{max}}]$ . We feed this vector to the morphology encoder \eqref{eq:morph-enc}:

\begin{equation}
z_m = f_{\psi}(m) = \text{ELU}\left( W_2\, \text{ELU}\left( W_1 m + b_1 \right) + b_2 \right) \in \mathbb{R}^{64}
\label{eq:morph-enc},
\end{equation}

where $W_1 \in \mathbb{R}^{128 \times d_m}$ and $W_2 \in \mathbb{R}^{64 \times 128}$ and the $z_{m}$, morphological embedding, if fed into the policy, for learning morph-conditioned policy.

\subsection{Policy Architecture}

In our policy architecture, we have divided inputs into i) regular observation, that are available during deployment and ii) privileged observations, that are not available during deployment and are  available in simulation only and finally iii) a sliding window observations, that need to be inferred during the deployed of the policy on the robot. So equation \ref{eq:teacher} uses privileged observation during training.
\begin{equation}
e_t = \sigma\left( W^{(e)} o_{priv} \right).
\label{eq:teacher}
\end{equation}
This equation is swapped by Equation \ref{eq:student}, from the adaptation module of our policy, and is used during the deployment of the policy:
\begin{equation}
l_t = g_{\alpha}(h_t) = \sigma\left( W^{(h)} h_t \right).
\label{eq:student}
\end{equation}

The input $x_t$ which takes the morphological embedding $z_{m}$ from the morphological encoder, which takes the morph vector, takes either the teacher (during training) or the student (during deployment) and if finally fed into the policy:
\begin{equation}
x_t = \left[\, o_t, \; e_t \ \text{(teacher)} \ \text{or} \ l_t \ \text{(student)}, \; z_m \, \right].
\label{eq:conc-teacher or stu}
\end{equation}

So both our equations , namely the actor policy \eqref{eq:actor policy} and value function  \ref{eq:critic policy or value function}, use the concatenated $x_t$ input.
\begin{equation}
\pi_{\theta}(a_t \mid o_t, h_t, m) = \mathcal{N}\left( \mu_t, \, \mathrm{diag}(\sigma^2) \right), \quad
\mu_t = f_{\theta_{\text{act}}}\left( [\, o_t, \kappa_t, z_m \,] \right)
\label{eq:actor policy}
\end{equation}
Overall, learning a morphology conditioned policy, where every forward pass is also condition on the randomized morphology, is selected from the range of morphologies (Table 1)as:
\begin{equation}
V_{\phi}(s_t, m) = f_{\phi_{\text{crt}}}\left( [\, o_t, \kappa_t, z_m \,] \right)
\label{eq:critic policy or value function}
\end{equation}

The above equations gives the morph-conditioned equation of our value function.

\subsection{Training Objective (PPO Optimization with morph-embedding)}

We optimize our morphology conditioned actor-critic, where morphology latent $z_m$ is an input to both of these actor policy and value function, using the PPO optimization. So, we are not modifying PPO inherently just conditioning it on the input $x_t$, which includes morphology latent.

\begin{equation}
\mathcal{L}^{\text{clip}}_t(\theta) = -\min \left( r_t(\theta) \, \hat{A}_t, \; \text{clip}\left( r_t(\theta), 1 - \epsilon, 1 + \epsilon \right) \, \hat{A}_t \right)
\label{eq:policy loss}
\end{equation}

The policy surrogate loss equation is given by \eqref{eq:policy loss}

\begin{equation}
\mathcal{L}^{\text{vf}}_t(\phi) = \frac{1}{2} \max\left( \left( V_{\phi}(x_t) - \hat{R}_t \right)^2, \; \left( V^{\text{old}}_{\phi}(x_t) + \text{clip}\left( V_{\phi}(x_t) - V^{\text{old}}_{\phi}(x_t), -\epsilon, \epsilon \right) - \hat{R}_t \right)^2 \right)
\label{eq:value loss}
\end{equation}

We trained our value function or critic loss based on regression loss, given by equation  \eqref{eq:value loss}
\begin{equation}
\mathcal{L}_{\text{PPO}}(\theta, \phi) = \frac{1}{|B|} \sum_{t \in B} \left[ \mathcal{L}^{\text{clip}}_t + c_v \, \mathcal{L}^{\text{vf}}_t - c_e \, \mathcal{S}_t \right]
\label{eq:total ppo loss}
\end{equation}

While the above PPO equation optimizes total PPO loss \eqref{eq:total ppo loss} . Overall clipping helps in achieving a robustness and stability.

\subsection{History-Aware Curriculum Learning (HACL)}
For curriculum, we expand the popular fixed-rule based \cite{margolis2024rapid} and game-inspired curriculum \cite{rudin2022learning} to incorporate history in the curriculum. We demonstrate using our experiments that a history-aware curriculum helps improved the overall learning pipeline for McARL.
   
\subsubsection{RNN predictions and Curriculum Binning}
To capture the history, we use Recurrent neural network, and extend the curriculum methodologies  \cite{margolis2024rapid}, \cite{rudin2022learning}, \cite{rudin2022advanced}, \cite{lee2020learning} which do not track or model history, but rather use rule-based method for increasing the difficulty. To address this and model the history or “hidden” state $H_{t-1}$ given by equation \eqref{eq:hidden state}, we utilize the a Recurrent Neural Net (RNN) based History-Aware Curriculum Learning (HACL) and integrate with our morphology conditioned learning McARL and highlighted the importance of combining it with McARL.

    
   

            

We use RNNs as they are better suited for capturing long-term dependencies owing to their  hidden state$H_{t-1}$, which captures the past hidden information perfectly and the curriculum can then adapt to the robot’s evolving performance. The curriculum bin selection is based on bin specific ID, and this input  $x_t$ \eqref{eq:one-hot} which is one-hot encoding of bins at time-step $t$:

\begin{equation}
x_t = \text{one-hot}(\text{BinID}_t) \in \mathbb{R}^{4000},
\label{eq:one-hot}
\end{equation}
where $4000$ (please refer to appendix tables) is the total number of bins and RNN predicts \eqref{eq:predictions} the expected linear and angular velocity for the encoded bins. For the next episode the bins with higher rewards are selected and as the data size increases, predictions gets more accurate and facilitate efficient bin selection and better learning. 

\begin{equation}
\hat{\mu}(b) = f_{\text{RNN}}(H_{t-1}, x_t) =
\begin{bmatrix}
\hat{r}_{\text{lin}}(b_i) \\
\hat{r}_{\text{ang}}(b_i)
\end{bmatrix},
\label{eq:predictions}
\end{equation}
where $H_{t-1}$ is the hidden state  \eqref{eq:hidden state} and $\hat{r}_{\text{lin}}(b_i)$, $\hat{r}_{\text{ang}}(b_i)$ are the predicted linear and angular rewards for the respective bin $b_i$.

\begin{equation}
H_t = \text{LSTM}(H_{t-1}, [x_t, r_t(b_t)], \theta).
\label{eq:hidden state}
\end{equation}
For our curriculum-network, we input the observed linear ($r_{lin}$) and angular ($r_{ang}$) reward ($r_t(b_t)$ and the bin IDs $\text{BinID}_t$ at time-step $t$.

\subsubsection{Curriculum weight update}
Each bin $(b_i)$in equation \eqref{eq:one-hot} is assigned a respective probability weight $(w_i)$ distribution given in equation $W = \{w_1, w_2, \dots, w_n\}$. The bins for which higher rewards predictions are made by the RNN predictions  \eqref{eq:predictions}, have their weights increased, which helps in sampling the commands from those bin and is given by the equation \eqref{eq:weight update}:


\begin{equation}
    w_{t+1}(b) = w_{t}(b) + \alpha \left(\hat{r}_{\text{lin}}(b) + \hat{r}_{\text{ang}}(b)\right),
\label{eq:weight update}
\end{equation}
As the training progresses, more data is provided by the robot's interaction with the environment in the form of observed linear ($r_{lin}$) and angular ($r_{ang}$) rewards, which in turn improve the RNN predictions and in turn also affect the command sampling for the next episode.

\begin{figure}[t]
    \centering
    \includegraphics[width=0.11\textwidth]{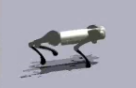}\hspace{1pt}
    \includegraphics[width=0.11\textwidth]{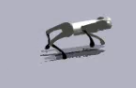}\hspace{1pt}
    \includegraphics[width=0.11\textwidth]{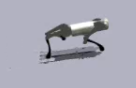}\hspace{1pt}
    \includegraphics[width=0.11\textwidth]{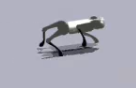}\hspace{1pt}
    \includegraphics[width=0.11\textwidth]{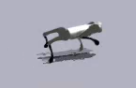}\hspace{1pt}
    \includegraphics[width=0.11\textwidth]{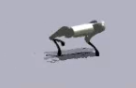}\hspace{1pt}
    \includegraphics[width=0.11\textwidth]{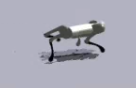}\hspace{1pt}
    \includegraphics[width=0.11\textwidth]{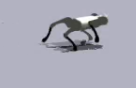}

    \vspace{2pt}

    \includegraphics[width=0.11\textwidth]{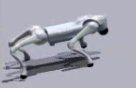}\hspace{1pt}
    \includegraphics[width=0.11\textwidth]{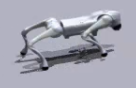}\hspace{1pt}
    \includegraphics[width=0.11\textwidth]{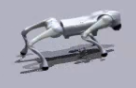}\hspace{1pt}
    \includegraphics[width=0.11\textwidth]{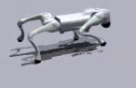}\hspace{1pt}
    \includegraphics[width=0.11\textwidth]{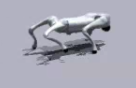}\hspace{1pt}
    \includegraphics[width=0.11\textwidth]{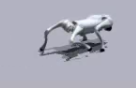}\hspace{1pt}
    \includegraphics[width=0.11\textwidth]{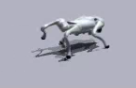}\hspace{1pt}
    \includegraphics[width=0.11\textwidth]{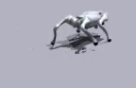}

     \vspace{2pt}
    \includegraphics[width=0.11\textwidth]{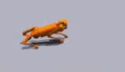}\hspace{1pt}
    \includegraphics[width=0.11\textwidth]{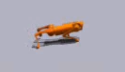}\hspace{1pt}
    \includegraphics[width=0.11\textwidth]{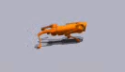}\hspace{1pt}
    \includegraphics[width=0.11\textwidth]{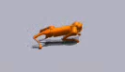}\hspace{1pt}
    \includegraphics[width=0.11\textwidth]{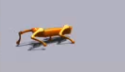}\hspace{1pt}
    \includegraphics[width=0.11\textwidth]{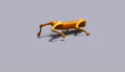}\hspace{1pt}
    \includegraphics[width=0.11\textwidth]{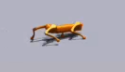}\hspace{1pt}
    \includegraphics[width=0.11\textwidth]{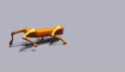}

    \vspace{2pt}

    \includegraphics[width=0.11\textwidth]{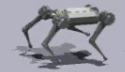}\hspace{1pt}
    \includegraphics[width=0.11\textwidth]{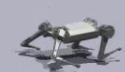}\hspace{1pt}
    \includegraphics[width=0.11\textwidth]{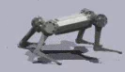}\hspace{1pt}
    \includegraphics[width=0.11\textwidth]{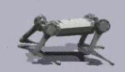}\hspace{1pt}
    \includegraphics[width=0.11\textwidth]{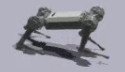}\hspace{1pt}
    \includegraphics[width=0.11\textwidth]{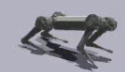}\hspace{1pt}
    \includegraphics[width=0.11\textwidth]{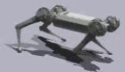}\hspace{1pt}
    \includegraphics[width=0.11\textwidth]{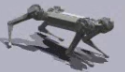}

    \vspace{2pt}

    \includegraphics[width=0.11\textwidth]{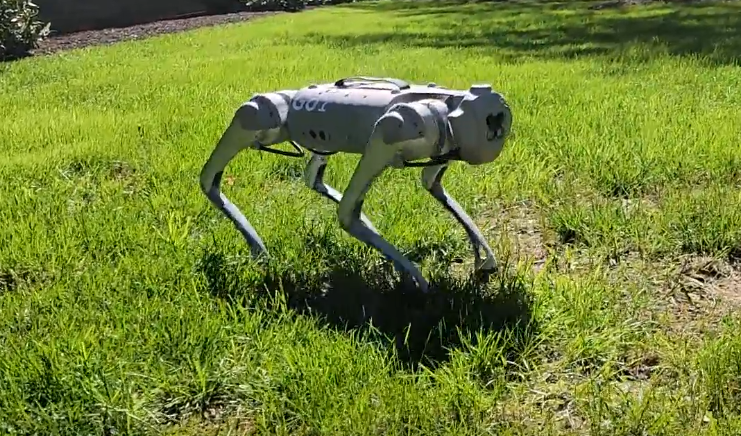}\hspace{1pt}
    \includegraphics[width=0.11\textwidth]{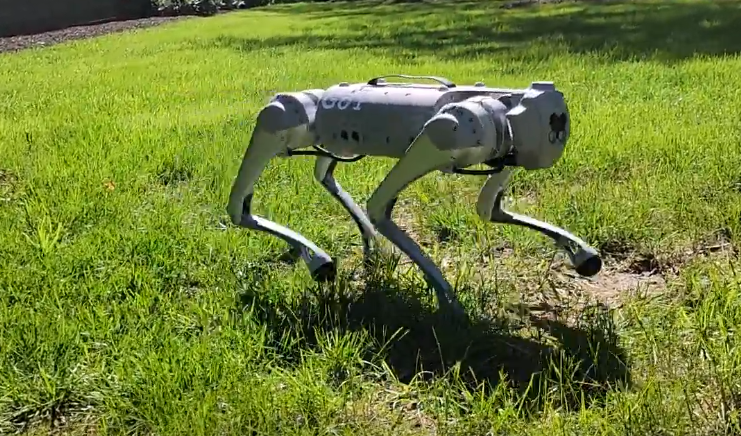}\hspace{1pt}
    \includegraphics[width=0.11\textwidth]{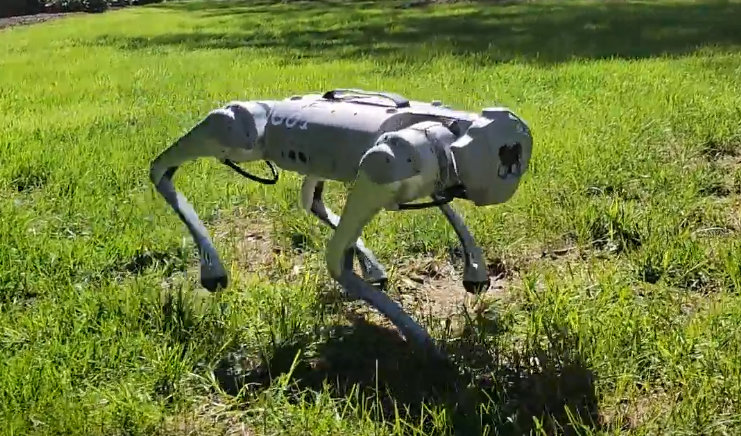}\hspace{1pt}
    \includegraphics[width=0.11\textwidth]{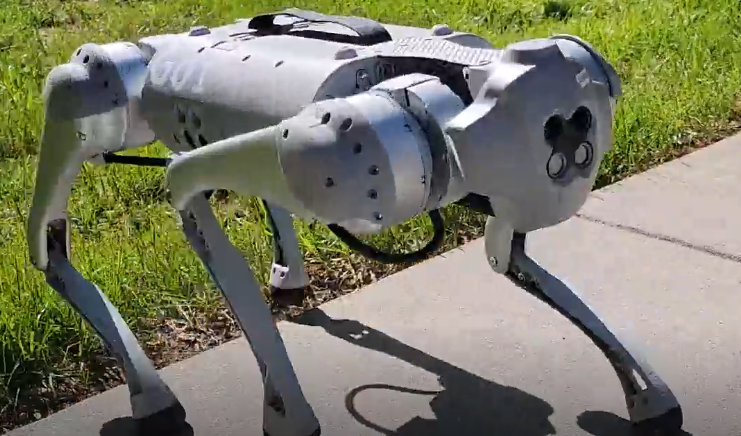}\hspace{1pt}
    \includegraphics[width=0.11\textwidth]{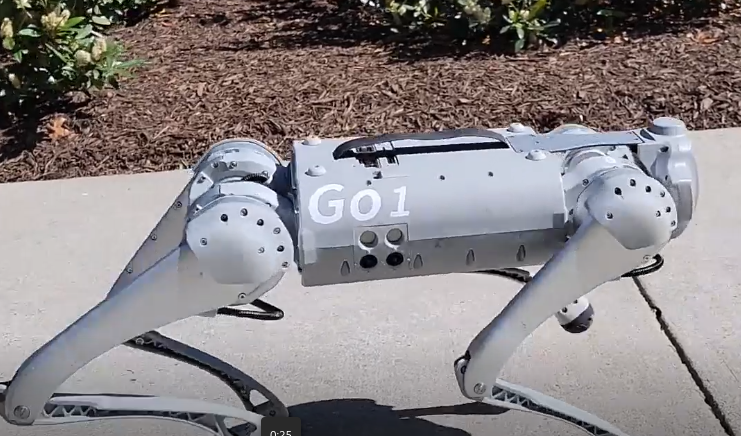}\hspace{1pt}
    \includegraphics[width=0.11\textwidth]{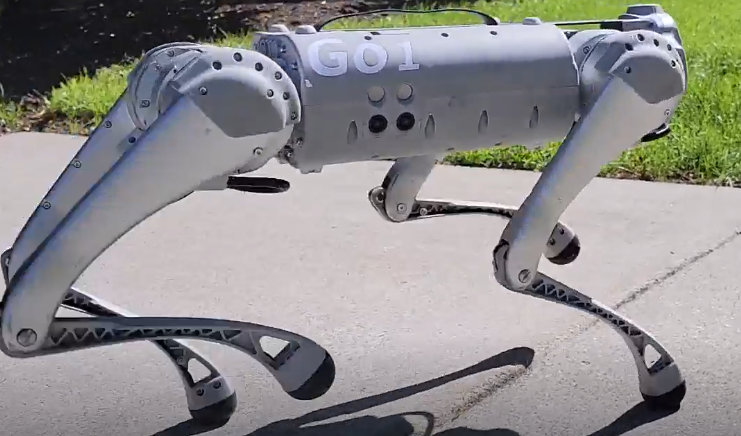}\hspace{1pt}
    \includegraphics[width=0.11\textwidth]{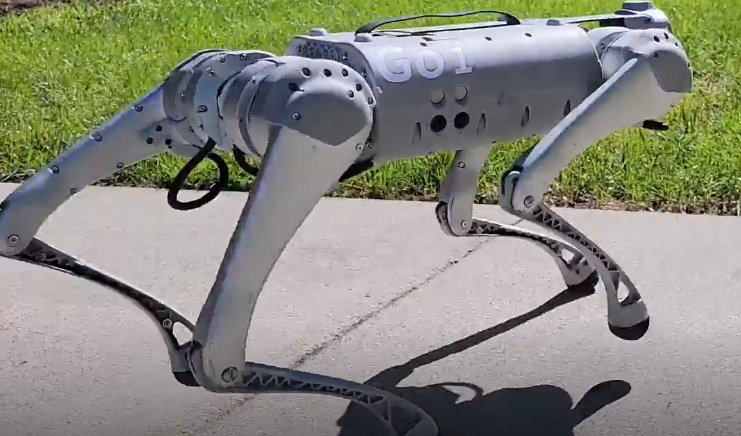}\hspace{1pt}
    \includegraphics[width=0.11\textwidth]{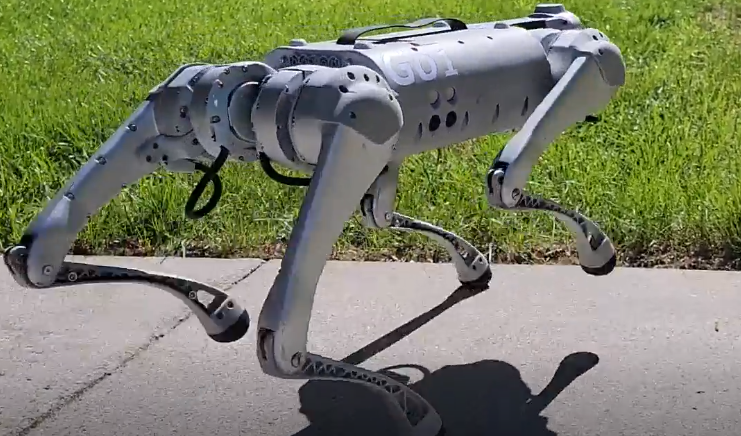}

    \caption{Morphology embedded policy controller testing in simulator on go1 robot (row 1) and Sim-to-sim zero shot transfer on go2, a1 and mini cheetah (row 2, 3, 4). We also perform real world testing on go1 robot for command velocity of 1m/sec (Last row).}
    \label{fig:images}
\end{figure}


\section{Experimental Setup}
\label{sec:setup}

We validate the effectiveness of McARL in both simulation and real-world and tested the validity of it for sim-to-real transfer and the sim-to-sim transfer. So we have trained McARL on Unitree Go1 robot and performed zero-shot transfer on go2, Mini-Cheetah and A1 robots. Also to check the robustness of our method we have tried training on other robot like go2, Mini-cheetah and A1 and then doing the transfer on the other remaining 3 robots. 


\subsection{Episode Design}

We have designed the episode to last for total of 20 seconds and timestep of 0.005 seconds and total of 4000 environments. An episode is terminated if it completes the time of robot fails, then episode ends. At the start of each episode, a new set of morphology vectors are generate for all the 4000 environments and most importantly each morphology vectors is separate instead of one vector for all environments, as that would act as a noise and the policy might not learn better. (Details of the episode are given in the appendix sections

\begin{table*}[ht]
\centering

\begin{tabular}{|l|l|c|}
\hline
\textbf{Reward Term} & \textbf{Equation } & \textbf{Weight} \\
\hline
\rowcolor{teal!10}
$r_{\text{lin}}$ (reward for tracking linear velocity) & $3 \cdot \exp\left(-\frac{||\vec{v}_{xy} - \vec{v}_{xy}^{\text{cmd}}||^2}{\sigma}\right)$ & 3.0 \\
\rowcolor{teal!10}
$r_{\text{ang}}$ (reward for tracking angular velocity) & $3 \cdot \exp\left(-\frac{||\vec{v}_{xy} - \vec{v}_{xy}^{\text{cmd}}||^2}{\sigma}\right)$ & 3.0 \\
\rowcolor{blue!10}
$r_{\text{feet\_air\_time}}$ & $\sum_{\text{feet}} (\text{air\_time} - 0.5) \cdot \mathbb{I}_{\text{contact}}$ & - \\
\rowcolor{blue!10}
$r_{\text{energy}}$ & $\sum_i \tau_i \cdot \dot{q}_i$ & - \\
\rowcolor{blue!10}
$r_{\text{energy\_efficiency}}$ & $- \left| \frac{\sum_i \tau_i \cdot \dot{q}_i}{||\vec{v}_{xy}|| \cdot \Delta t + \epsilon} \right|$ & $1\cdot 10^{-7}$ \\
\rowcolor{orange!10}
$r_{\text{lin\_vel\_z}}$ & $v_z^2$ & - \\
\rowcolor{orange!10}
$r_{\text{ang\_vel\_xy}}$ & $||\vec{\omega}_{xy}||^2$ & - \\
\rowcolor{orange!10}
$r_{\text{orientation}}$ & $||g_{xy}||^2$ & - \\
\rowcolor{orange!10}
$r_{\text{base\_height}}$ & $2(h - h^{target})^2$ & - \\
\rowcolor{orange!10}
$r_{\text{torques}}$ & $||\vec{\tau}||^2$ & - \\
\rowcolor{orange!10}
$r_{\text{dof\_vel}}$ & $||\dot{q}||^2$ & - \\
\rowcolor{orange!10}
$r_{\text{dof\_acc}}$ & $||(\dot{q}_{t} - \dot{q}_{t-1}) / \Delta t||^2$ & - \\
\rowcolor{orange!10}
$r_{\text{action\_rate}}$ & $0.2 \cdot ||a_t - a_{t-1}||^2$ & 0.2 \\
\rowcolor{orange!10}
$r_{\text{collision}}$ & $\sum_{\text{contact}} \mathbb{I}_{||F|| > 0.1}$ & - \\
\rowcolor{orange!10}
$r_{\text{termination}}$ & $\mathbb{I}_{\text{reset}} \cdot \neg \mathbb{I}_{\text{timeout}}$ & - \\
\rowcolor{orange!10}
$r_{\text{survival}}$ & $\neg(\mathbb{I}_{\text{reset}} \cdot \neg \mathbb{I}_{\text{timeout}})$ & - \\
\rowcolor{orange!10}
$r_{\text{dof\_pos\_limits}}$ & $\sum_i \max(0, q_{i}^{min} - q_i, q_i - q_i^{max})$ & - \\
\rowcolor{orange!10}
$r_{\text{dof\_vel\_limits}}$ & $\sum_i \text{clip}(|\dot{q}_i| - \dot{q}_i^{max}, 0, 1)$ & - \\
\rowcolor{orange!10}
$r_{\text{torque\_limits}}$ & $\sum_i \max(0, |\tau_i| - \tau_i^{max})$ & - \\
\rowcolor{orange!10}
$r_{\text{stand\_still}}$ & $\sum_i |q_i - q_i^{default}| \cdot \mathbb{I}_{||\vec{v}_{cmd}|| < 0.1}$ & - \\
\rowcolor{orange!10}
$r_{\text{stumble}}$ & $\mathbb{I}_{||F_{xy}^{foot}|| > 5 \cdot |F_z^{foot}|}$ & - \\
\rowcolor{orange!10}
$r_{\text{feet\_contact\_forces}}$ & $\sum_{\text{feet}} \max(0, ||F|| - F^{max})$ & - \\
\hline
\end{tabular}
\caption{\textbf{Reward structure: } An overview of our reward strutures and tasks for our \textbf{McARL framework }\textcolor{teal}{task rewards}, the main function of these rewards is to accruately track the commanded linear and angular velocity \textcolor{blue}{augmented auxiliary rewards}, these rewards promote the efficient and dynamic motion and are neccessary to maintain stable gait, and \textcolor{orange}{fixed auxiliary rewards}, these rewards help in maintaining a realistic and dynamic motion.}
\label{tab:reward_structure}
\end{table*}

\subsection{Simulation details \& Hyperparameter details}
We simulate Unitree Go1, Go2, Mini Cheetah and A1 robots for McARL evaluation, using these robot's respective URDF files in the simulated environment(IsaacGym Simulator\cite{makoviychuk2021isaac}). Each RL episode of simulation lasts for roughly 20s with a total of 4000 parallel environments with a time-step of $dt$ = 0.005. And the sampled command velocities are provided at the start or each episode. We have adapted our code mostly from the open-source repositories\cite{margolis2024rapid},\cite{rudin2022learning}. Similar to \cite{margolis2024rapid}, we have also trained 400 million timestep in simulation using 4000 environments of A1, Go1, Go2, and MIT mini cheetah robots for validating McARL. For training purpose, we have used an Nvidia RTX 4090 laptop-based GPU, completing the full simulation within less than 2 hours. (Pleas refer to parameters details in appendix)



\subsection{Baselines \& Evaluation metrics}

We have compared McARL with the current state of art work in legged robots, for both morphology-aware and morphology-unaware. We evaluated against a set of parameters like morphology generalization, maximum speed, number of robots and the real-world tests.

\begin{table}[ht]
\centering
\caption{Comparison of morphology generalization and real-world testing across methods.}
\label{tab:method-comparison}
\resizebox{\linewidth}{!}{
\begin{tabular}{l c c c l}
\toprule
\textbf{Method} & \textbf{Morph Generalization} & \textbf{Real-World Test} & \textbf{Max Speed (m/s)} & \textbf{Robots Used} \\
\midrule
GenLoco (CoRL '22)   & Yes (sim $\rightarrow$ sim \& real) & Yes (MC, A1) & $\sim$3.1         & MC, A1 \\
BodyTransfomer  & Yes (sim $\rightarrow$ sim \& real) & Yes (MC, A1) & $\sim$3.1         & MC, A1 \\
ManyQuadrupedal  & Yes (sim $\rightarrow$ sim \& real) & Yes (MC, A1) & $\sim$3.1         & MC, A1 \\
One Policy to run them all  & Yes (sim $\rightarrow$ sim \& real) & Yes (MC, A1) & $\sim$3.1         & MC, A1 \\
MetaMorph         & No (1 robot)                        & Partial sim $\rightarrow$ real & 1.5–2.0         & A1 \\
DMAP           & No (1 robot)                        & Yes          & $\sim$1.0         & Cassie only \\
\textbf{Ours (McARL)} & \textbf{Yes (Go1 $\rightarrow$ Go2, MC, A1)} & \textbf{Yes (Go1 real)} & \textbf{6.0 (Go1), 3.5 (Go2)} & Go1, Go2, A1, MC \\
\bottomrule
\end{tabular}
}
\end{table}

To our knowledge, McARL outperform all the SOTA work as evident by the table above and achieves much better zero-shot transfer speed for different morphologies and the unique feature of our method is that it is trained using only one robot morphology and explicitly conditioning the actor-critic network using randomized morphology latent.


\section{Results}
\label{sec:results}

In this section, we compare the results of our McARL in both sim-to-sim transfer and also sim-to-real transfer. We compare our policy with various training strategies and demonstrated the importance or morphology in policy learning and also the challenges encountered during the testing.

\begin{table*}[ht]
\centering
\vspace{-0.5em}
\begin{minipage}[t]{0.48\textwidth}
\vspace*{0pt}
\centering
\caption*{(a) All the important PPO Hyperparameters used for training McARL}
\begin{tabular}{|l|l|}
\hline
\textbf{Parameter} & \textbf{Value} \\
\hline
Value Loss Coefficient & 1.0 \\
Use Clipped Value Loss & True \\
Clipping Parameter & 0.2 \\
Entropy Coefficient & 0.01 \\
Learning Rate & $1 \times 10^{-3}$ \\
Adaptation Module Learning Rate & $1 \times 10^{-3}$ \\
Num PPO Epochs & 5 \\
Num Mini Batches & 4 \\
Schedule & Adaptive \\
Discount Factor ($\gamma$) & 0.99 \\
GAE Lambda ($\lambda$) & 0.95 \\
Desired KL & 0.01 \\
Max Gradient Norm & 1.0 \\
Total Environments & 4000 \\
Optimizer & Adam \\
Total Timesteps & 400M \\
\hline
\end{tabular}
\end{minipage}
\hfill
\begin{minipage}[t]{0.48\textwidth}
\vspace*{0pt}
\centering
\caption*{(b) History-aware Curriculum Learning (HACL) Parameters}
\begin{tabular}{|l|l|}
\hline
\textbf{Parameter} & \textbf{Value} \\
\hline
$v_x^{\text{cmd}}$ initial & $[-1.0, 1.0]$ \\
$\omega_z^{\text{cmd}}$ initial & $[-1.0, 1.0]$ \\
$v_x^{\text{cmd}}$ max & $[-6.0, 6.0]$ \\
$\omega_z^{\text{cmd}}$ max & $[-5.0, 5.0]$ \\
$v_x^{\text{cmd}}$ bin size & 0.5 \\
$\omega_z^{\text{cmd}}$ bin size & 0.5 \\
Design Space (x, y, z) & (20, 10, 20) \\
Total Bins & 4000 \\
RNN Input Size & 4000 \\
RNN Hidden Size & 128 \\
RNN Output Size & 8000 \\
RNN Layers & 1 \\
Loss Function & MSELoss \\
Optimizer & Adam \\
Learning Rate & 0.001 \\
Batch Size & Variable (e.g., 10,000) \\
Output Prediction & $r_{\text{lin}}, r_{\text{ang}}$ \\
Weight Update Rule & $0.2 \times \frac{\hat{r}_{\text{lin}} + \hat{r}_{\text{ang}}}{2}$ \\
\hline
\end{tabular}
\end{minipage}

\vspace{1em}

\begin{minipage}[t]{0.4\textwidth}
\vspace*{0pt}
\centering
\caption*{(c) Actor-Critic Network architecture parameters used for McARL}
\begin{tabular}{|l|l|}
\hline
\textbf{Parameter} & \textbf{Value} \\
\hline
Actor Hidden Layers & [512, 256, 128] \\
Critic Hidden Layers & [512, 256, 128] \\
Activation Function & ELU \\
Adaptation Module Hidden Layers & [[256, 32]] \\
Env-Factor Encoder Input Dim & 18 \\
Env-Factor Encoder Latent Dim & 18 \\
Env-Factor Encoder Hidden Layers & [[256, 128]] \\
\hline
\end{tabular}
\end{minipage}
\hfill
\begin{minipage}[t]{0.4\textwidth}
\vspace*{0pt}
\centering
\caption*{(c) Morphology Encoder ($z_m$) Parameters used during McARL traiing}
\begin{tabular}{|l|l|}
\hline
\textbf{Parameter} & \textbf{Value} \\
\hline
Input Dimension & 14 \\
First Layer & 128 \\
First Activation Function & ELU \\
Second Layer & 64 \\
Output Latent Dimension & 64 \\
Second Activation Function & ELU \\
\hline
\end{tabular}
\end{minipage}
\caption{A brief overview of all the important parameters and hyperparameters that are used to train McARCL, which includes details of the PPO Hyperparameters and History-aware Curriculum Learning (HACL) Parameters (first row) and the Actor-critic network and Morphology Encoder ($z_m$) Parameters (bottom row) }
\label{tab:all_hyperparams}
\end{table*}


\subsection{Zero-Shot transfer}

We performed zero-shot transfer of our policy trained on go1 robot and transferred it to go2, mini cheetah and a1 robots. We compared different combination of PPO with morphology, non-morphology and various curriculum strategies and the results are elaborated in the table 3.

\begin{table}[ht]
\centering
\caption{Comparison of PPO in combination with morphology and control conditioned training and various input and curriculum learning strategies across robot morphologies. Mean forward speed (m/s) with standard deviation.}
\label{tab:robot-comparison}
\resizebox{\linewidth}{!}{
\begin{tabular}{l l l c c c c}
\toprule
\textbf{PPO Variants} & \textbf{Inputs} & \textbf{Curriculum} & \textbf{Go1 (Train)} & \textbf{Go2 (Zero-shot)} & \textbf{Mini Cheetah} & \textbf{A1} \\
\midrule
P0  & state                   & HACL     & 5.91 $\pm$ 0.1 & 1.36 $\pm$ 0.2 & 0.0 $\pm$ 0.0 & 0.0 $\pm$ 0.0 \\
P1  & state + ID (1 vec)             & Non-History      & 6.05 $\pm$ 0.13          & 3.42 $\pm$ 0.15 & 0.36 $\pm$ 0.2 & 0.0 $\pm$ 0.0 \\
P2  & state + morph           & Non-History     & 5.25 $\pm$ 0.15          & 2.87 $\pm$ 0.7 & 1.02 $\pm$ 0.2 & 0.34 $\pm$ 0.1 \\
P3  & state + morph           & HACL  & 5.98 $\pm$ 0.2          & 3.43 $\pm$ 0.2 & 1.47 $\pm$ 0.46 & 0.5 $\pm$ 0.27 \\
P4  & state + morph + ctrl    & HACL  & 6.0 $\pm$ 0.18 & 3.47 $\pm$ 0.1 & 0.5 $\pm$ 0.45 & 0.35 $\pm$ 0.1 \\
P5  & state + ctrl            & HACL  & 5.71 $\pm$ 0.1          & 3.4 $\pm$ 0.25 & 1.18 $\pm$ 0.5 & 0.3 $\pm$ 0.2 \\
\bottomrule
\end{tabular}
}
\end{table}

\subsection{Ablation Studies}

\textbf{Effect of morphology embedding vs state only:}
As its evident by table 3 above that, a policy conditioned on morphology embedding performs much better that non-morph embeddings policies. What is interesting was the result of the or P4 i.e policies which utilizes the vector with both morphology and controls parameter, one possible reason would be that stiffness and damping parameter, along with other parameter made the policy a bit conservative in taking stronger torques, which is why we saw a slight decreases in the performance.

\textbf{Curriculum Vs History-Aware Curriculum Vs No Curriculum:}
Similarly a history aware curriculum like HACL is a much better combination for morphology conditioned policy learning compared to fixed-rule based updated or even no curriculum approach. One of the reasons why HACL combination works better is due to the non-markovian nature of legged locomotion and apart from that if learn to predict some correlation between command velocities and the linear and angular rewards received, improved the overall locomotion stability and performance.

\begin{figure}[t]
    \centering
    \includegraphics[width=0.24\textwidth]{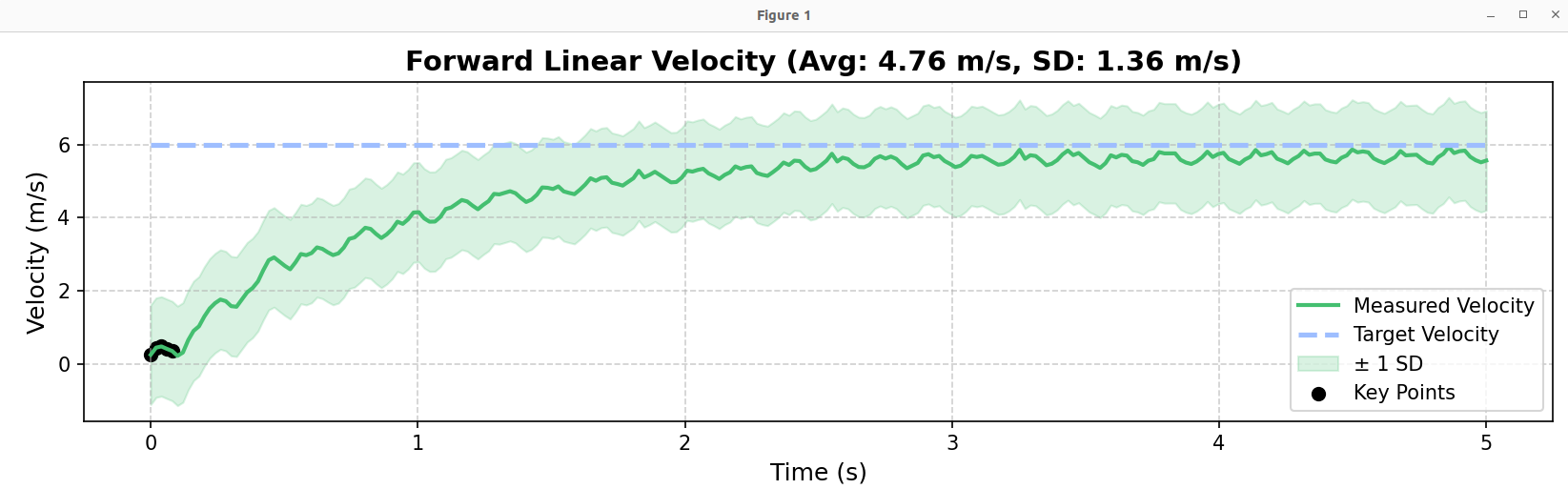}\hspace{1pt}
    \includegraphics[width=0.24\textwidth]{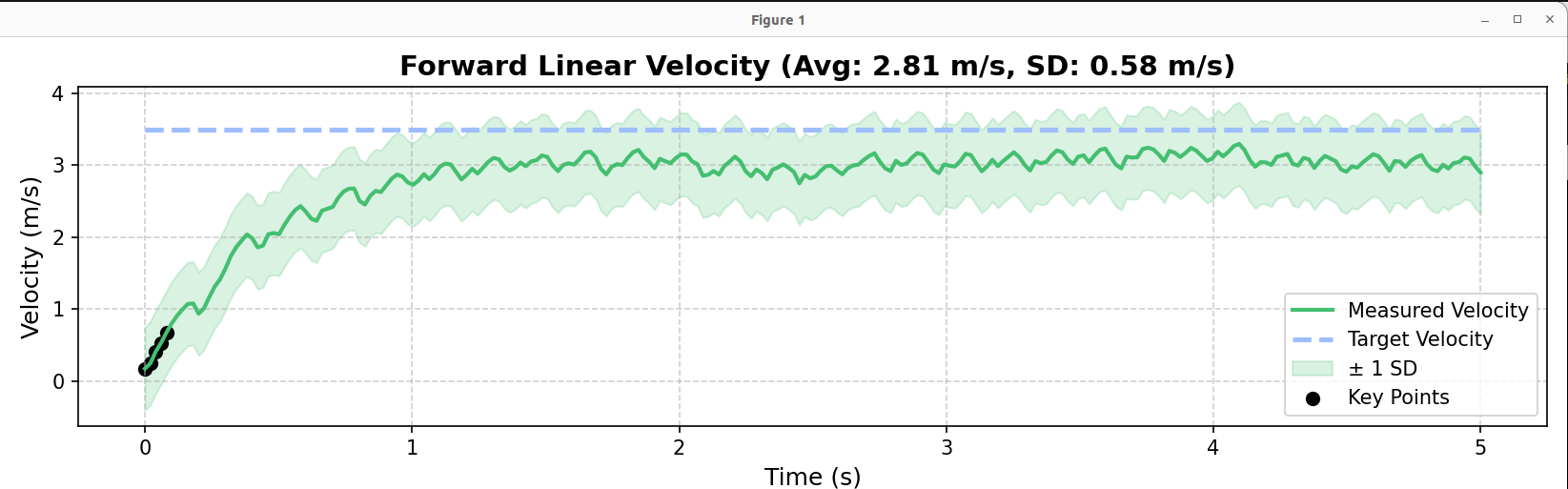}\hspace{1pt}
    \includegraphics[width=0.24\textwidth]{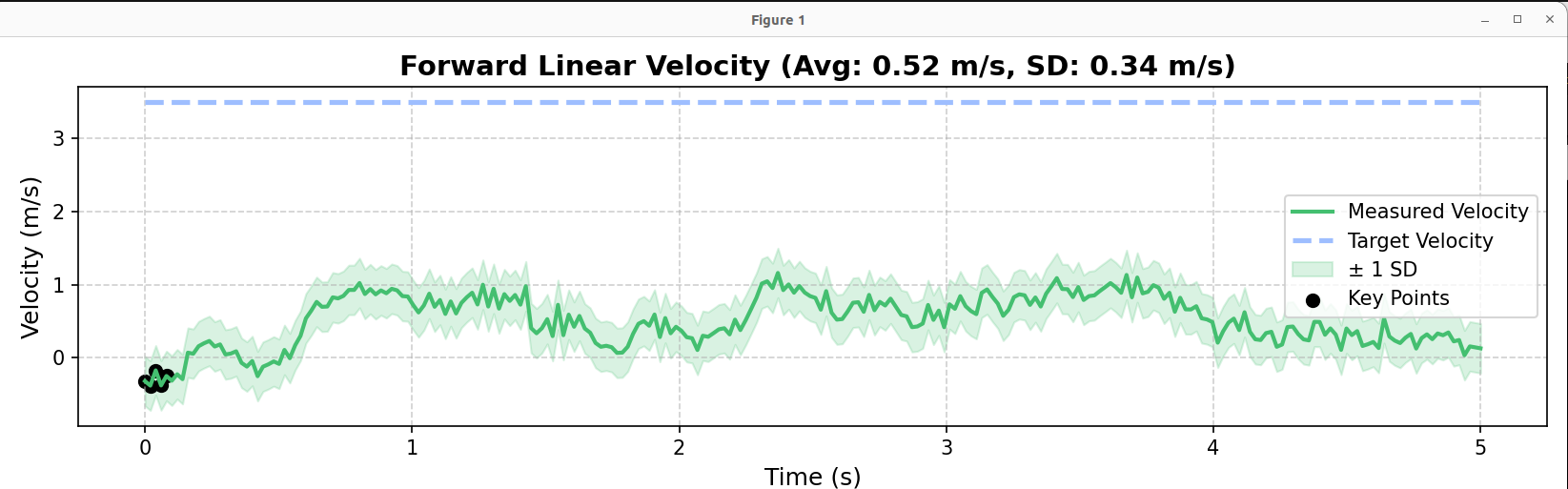}\hspace{1pt}
    \includegraphics[width=0.24\textwidth]{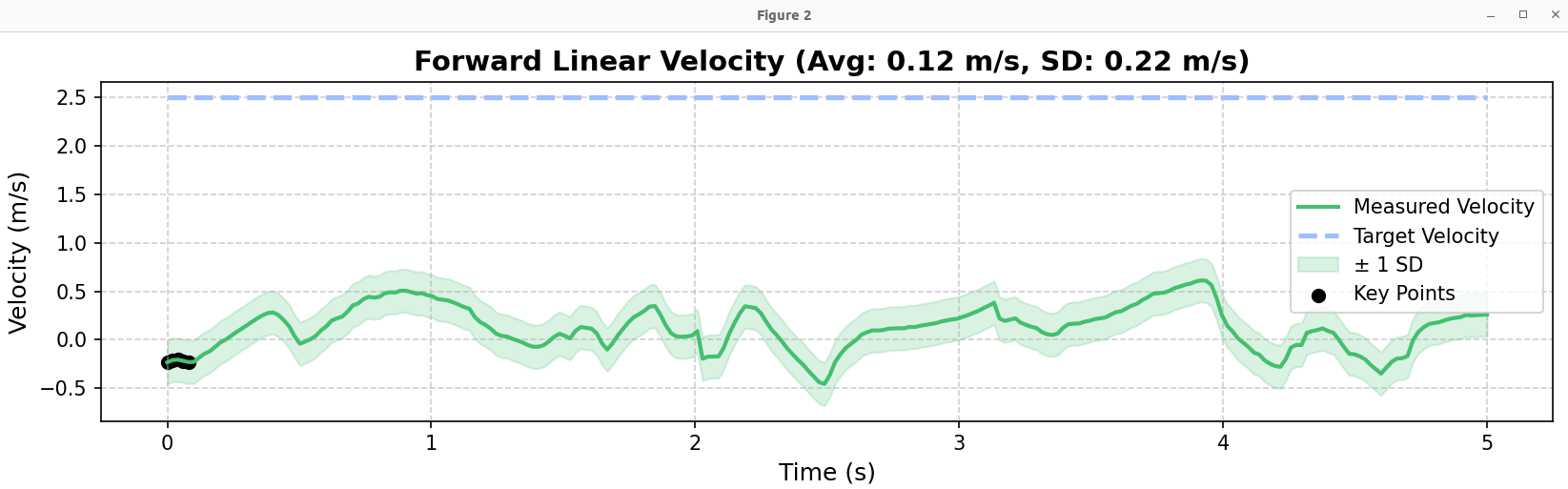}\hspace{1pt}
    
    \vspace{2pt}

     \includegraphics[width=0.24\textwidth]{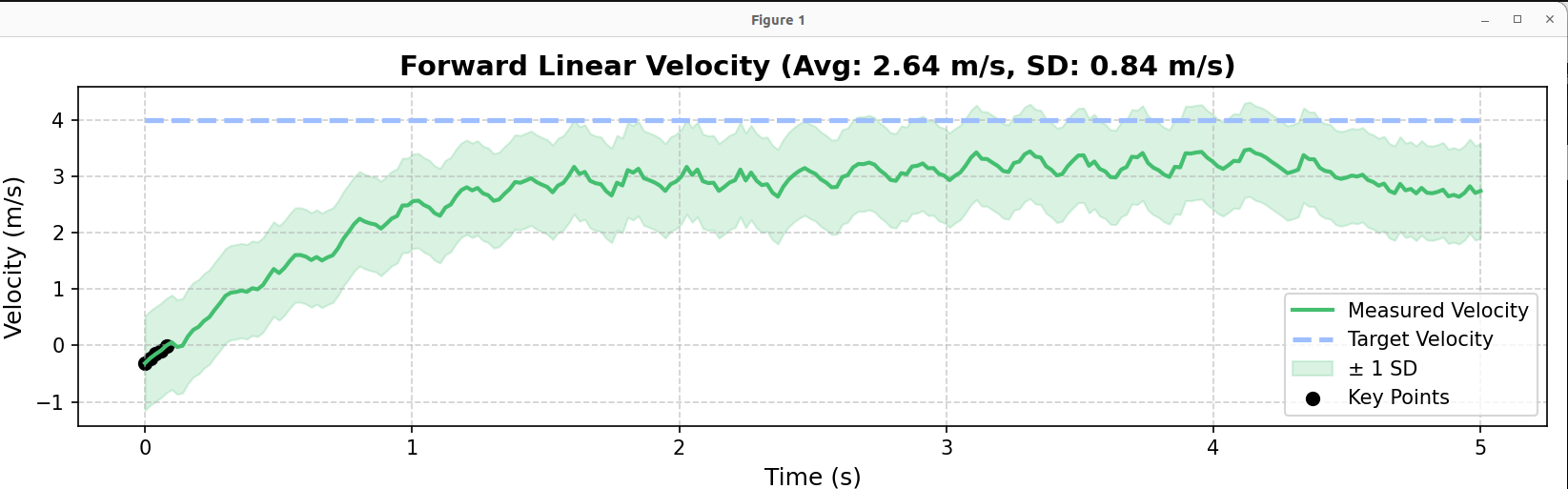}\hspace{1pt}
    \includegraphics[width=0.24\textwidth]{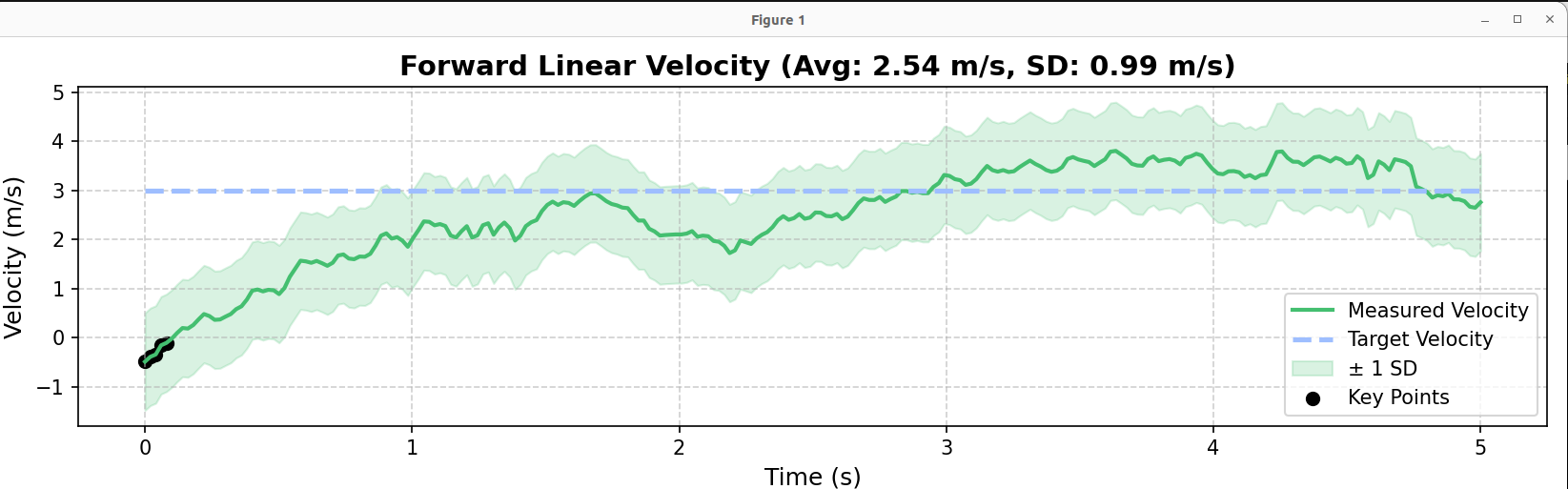}\hspace{1pt}
    \includegraphics[width=0.24\textwidth]{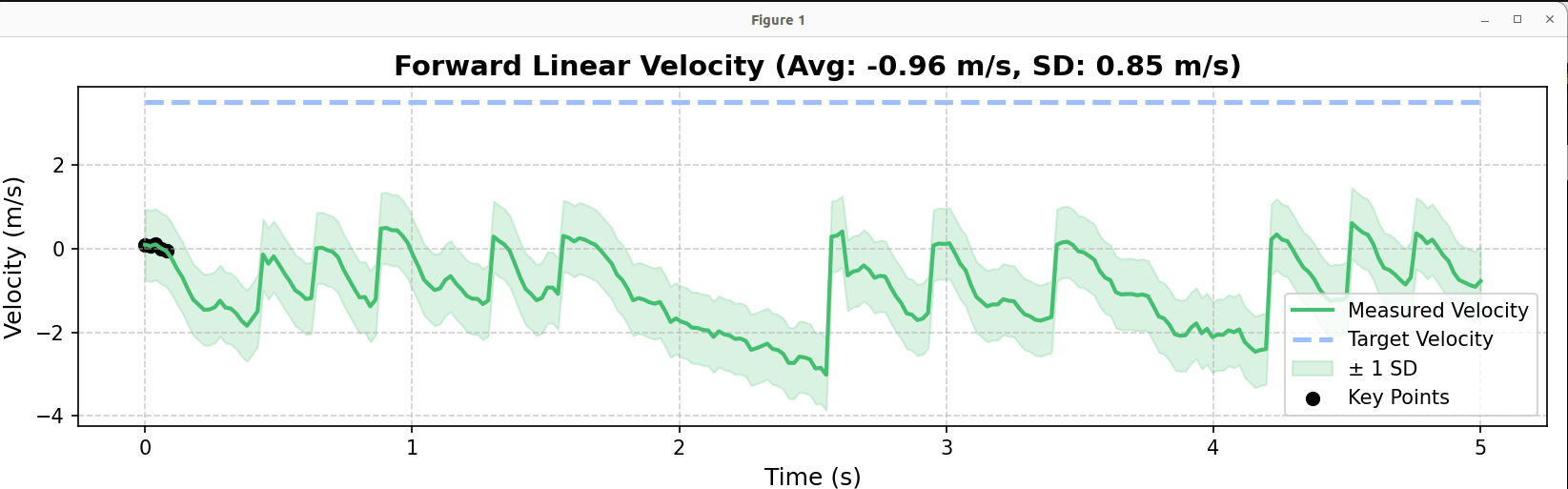}\hspace{1pt}
    \includegraphics[width=0.24\textwidth]{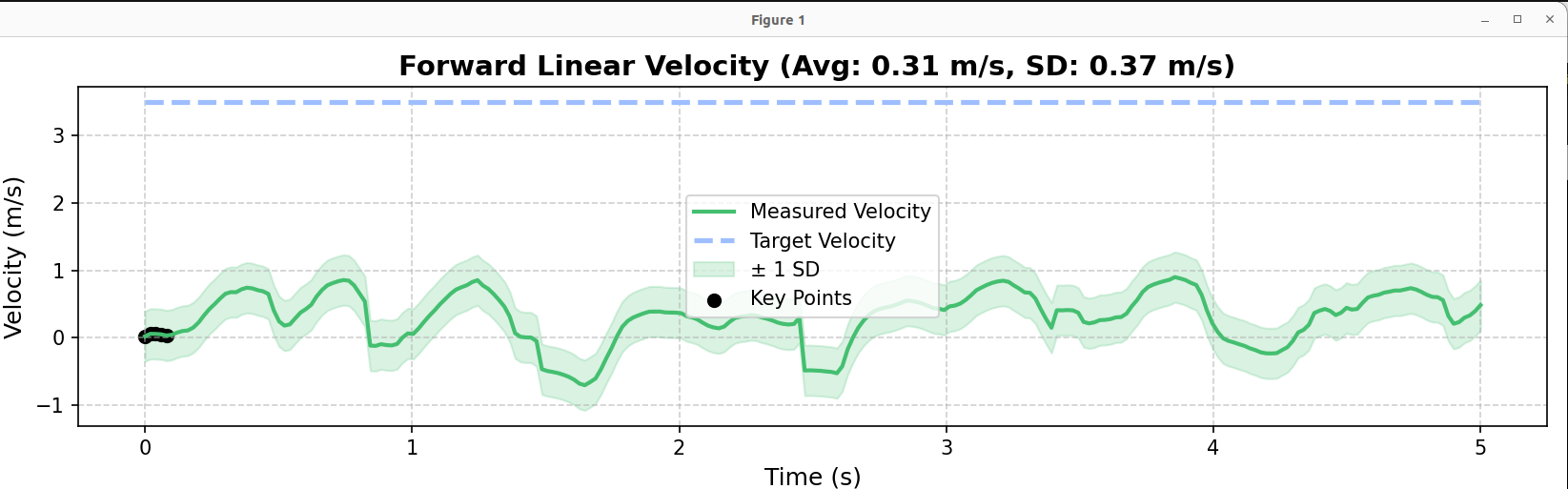}\hspace{1pt}

     \vspace{2pt}

     \includegraphics[width=0.24\textwidth]{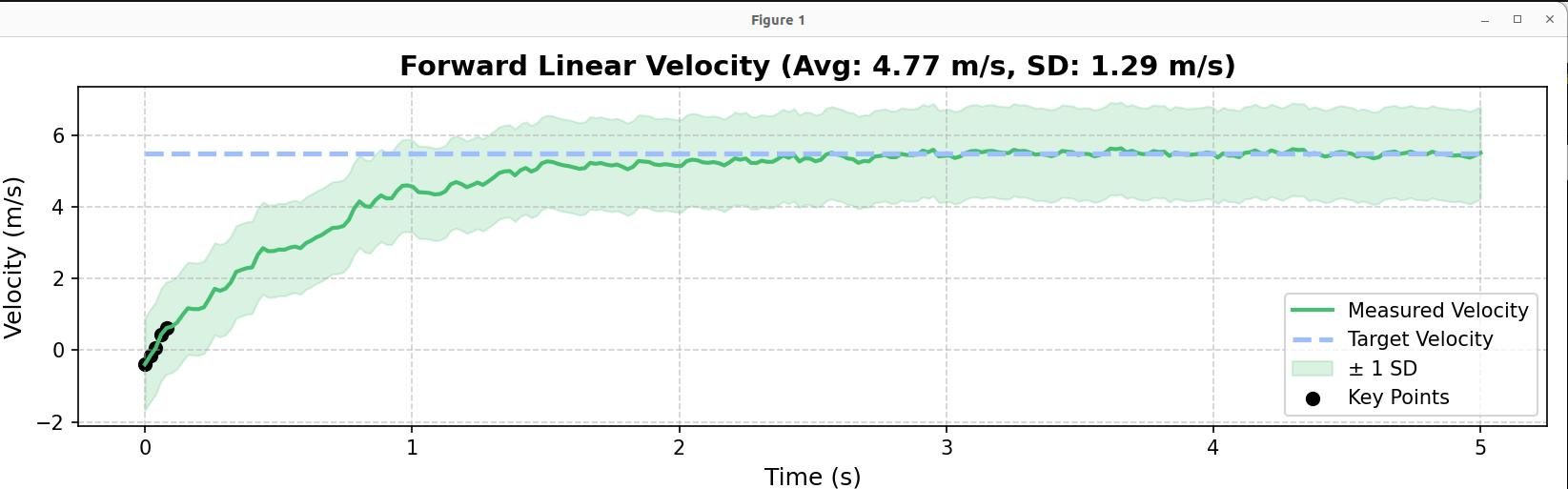}\hspace{1pt}
    \includegraphics[width=0.24\textwidth]{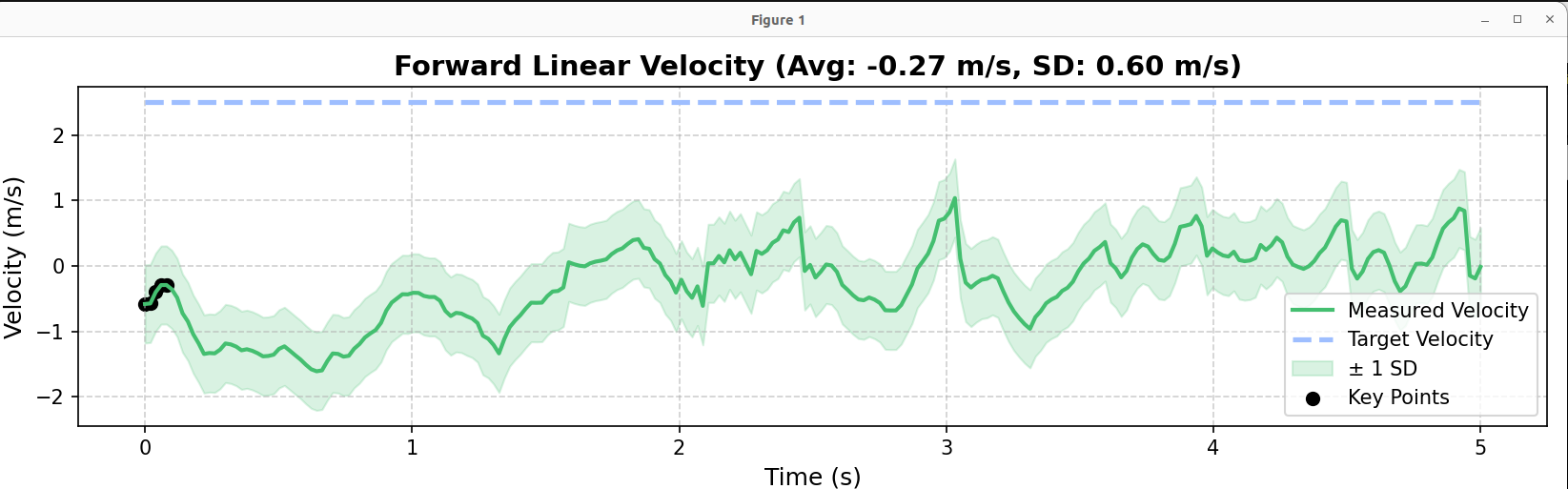}\hspace{1pt}
    \includegraphics[width=0.24\textwidth]{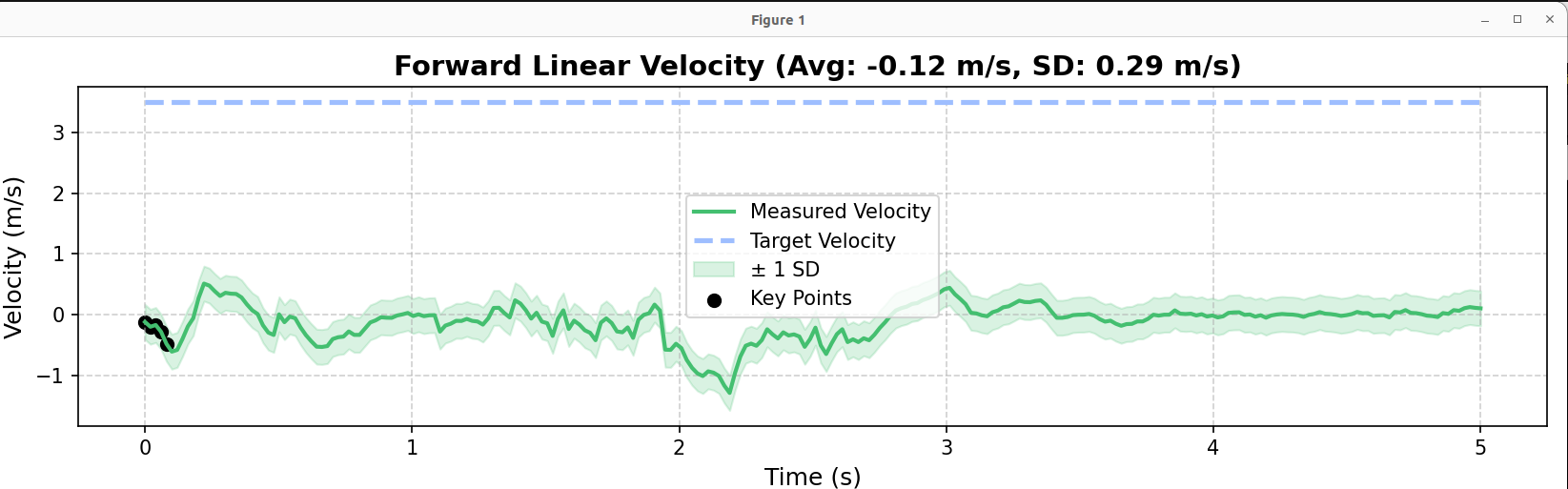}\hspace{1pt}
    \includegraphics[width=0.24\textwidth]{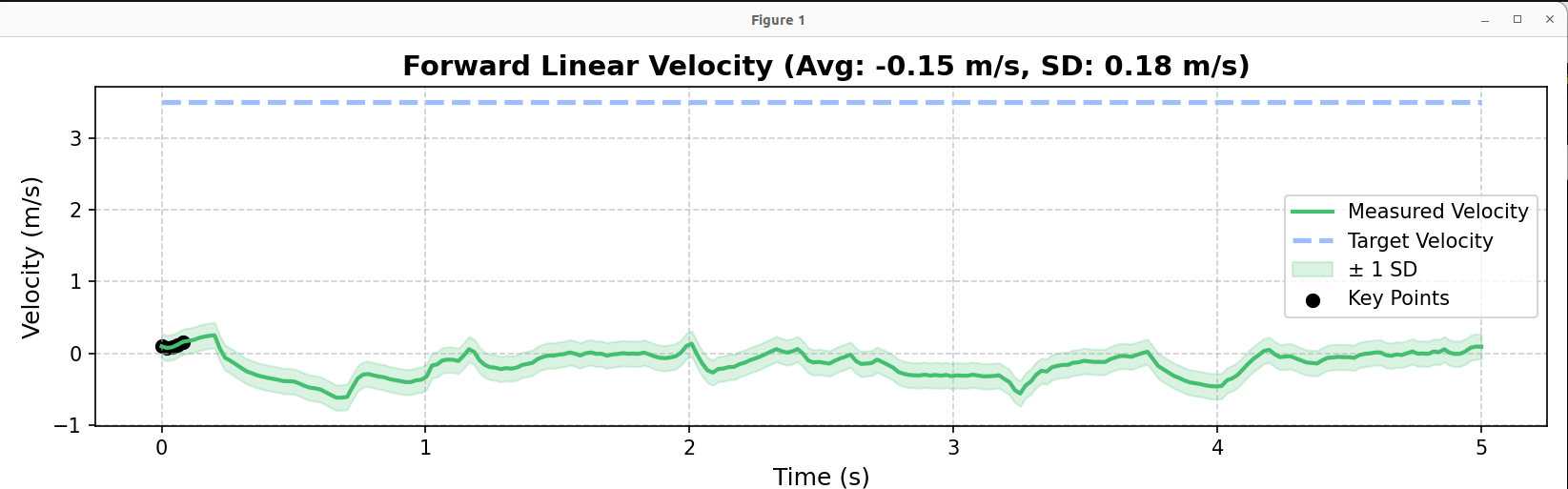}\hspace{1pt}
    
     \vspace{2pt}
     \includegraphics[width=0.24\textwidth]{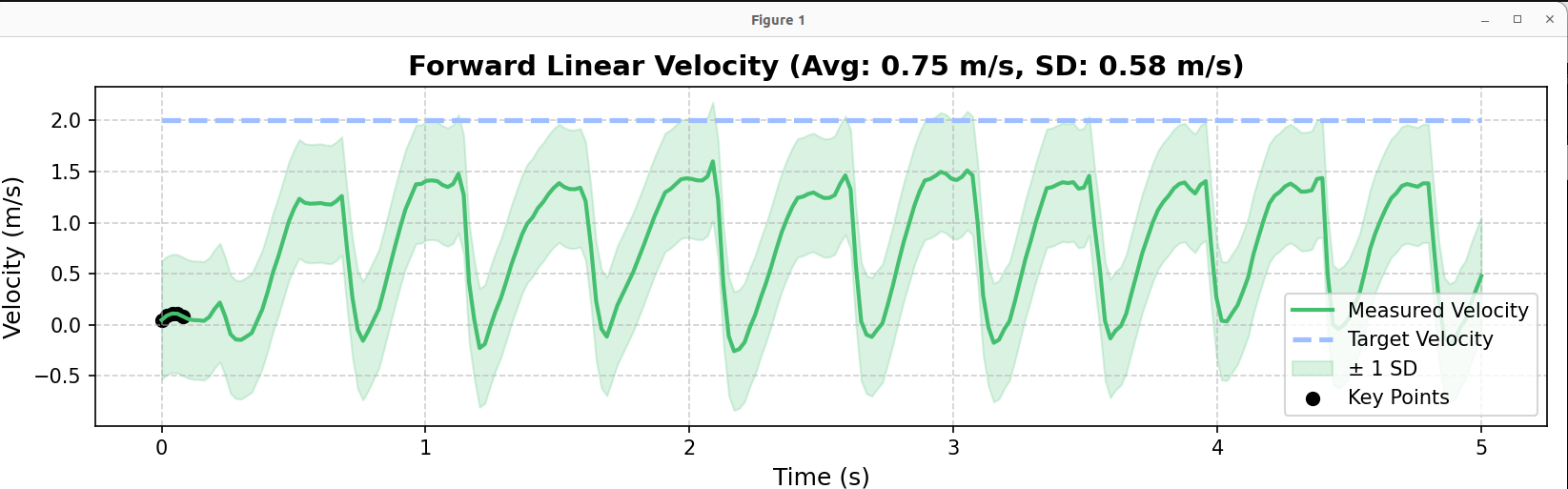}\hspace{1pt}
    \includegraphics[width=0.24\textwidth]{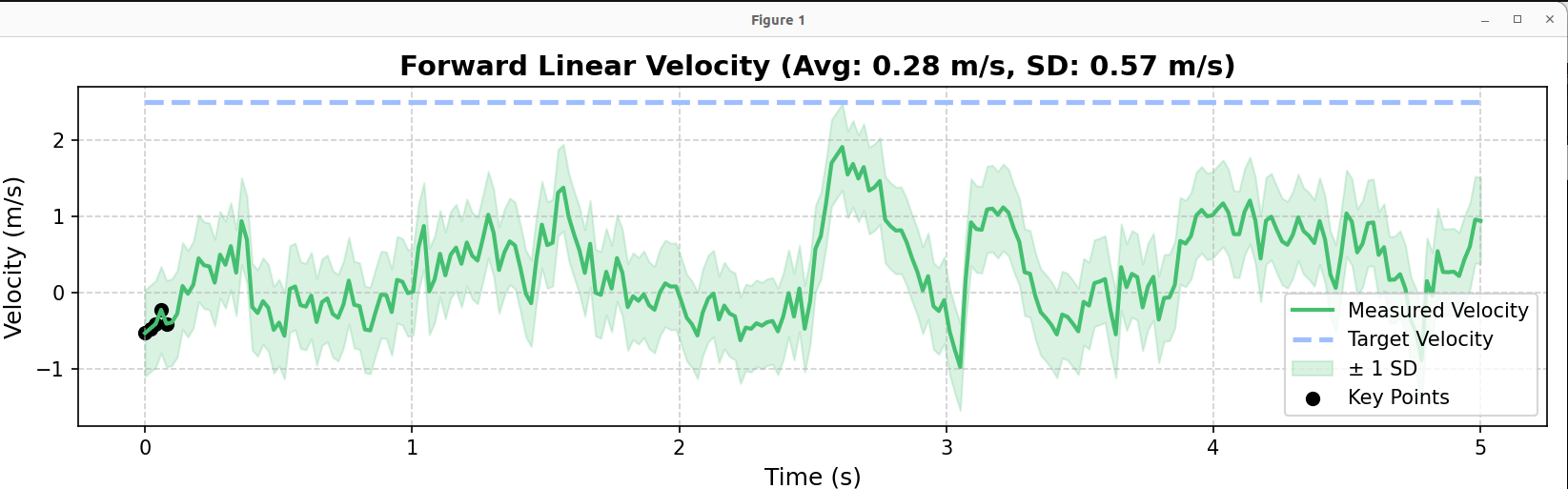}\hspace{1pt}
    \includegraphics[width=0.24\textwidth]{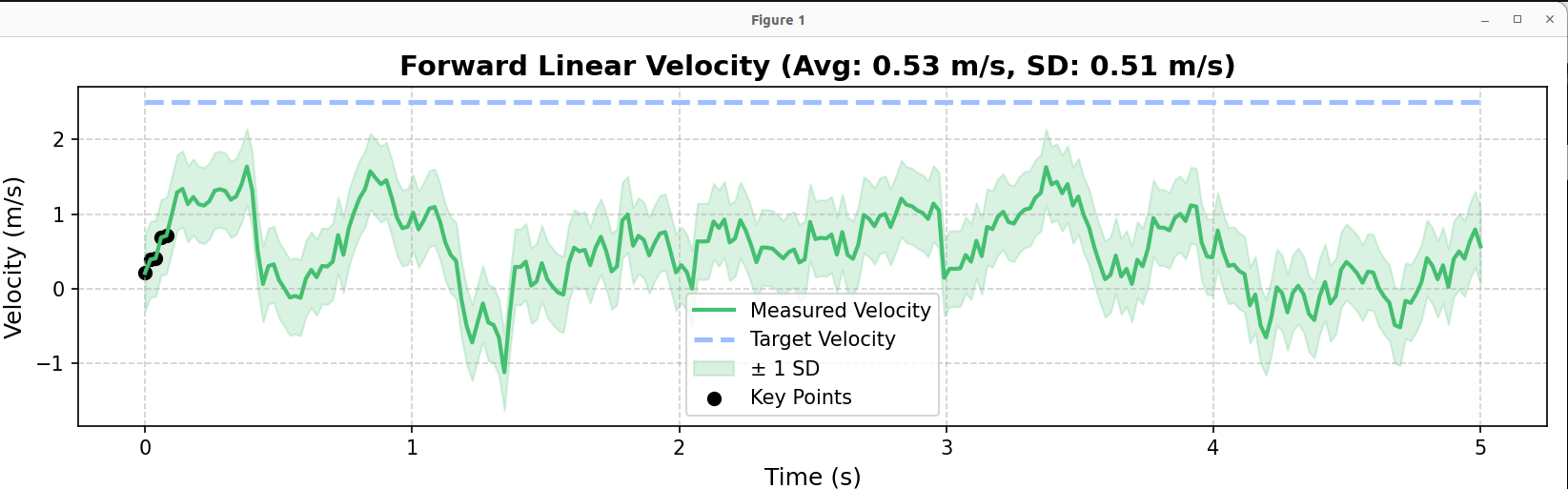}\hspace{1pt}
   \includegraphics[width=0.24\textwidth]{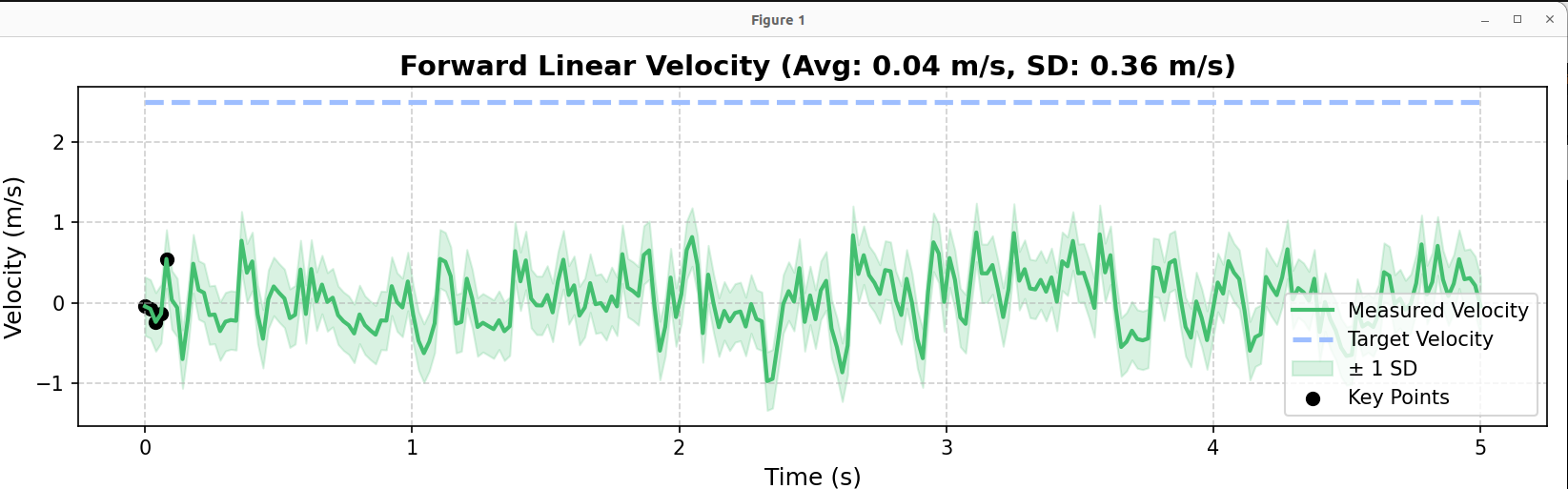}\hspace{1pt}

    \caption{The first image of each row represents on which ronot McARL has been trained and the remaining following images represent the transfer velocities on the other robots. The Go1 robot (first row) achieves great velocity of 6m/s with max transfer on go2 at around 3.5m/s, the Go2 (second row) robot reaches 3.5 m/s and similar transfer rate for go1, while the transfer loss starts to get bad  from Mini Cheetah robot (third row) and worse for the A1 robot (last row). Interstingly the A1 robots performance was not that stable and more of galloping behavior indicating the need for fine tuning the config parameters and tuning the rewards accordingly.}
    \label{fig:two_row_eight_images1}
\end{figure}

\begin{figure}[t]
    \centering
    \includegraphics[width=0.2434\textwidth]{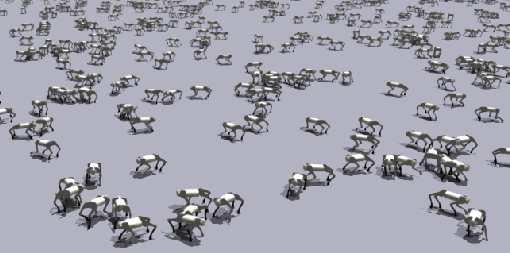}\hspace{1pt}
    \includegraphics[width=0.2434\textwidth]{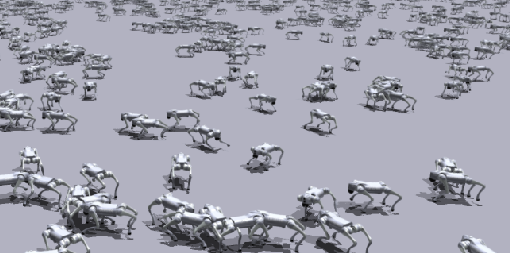}\hspace{1pt}
    \includegraphics[width=0.2434\textwidth]{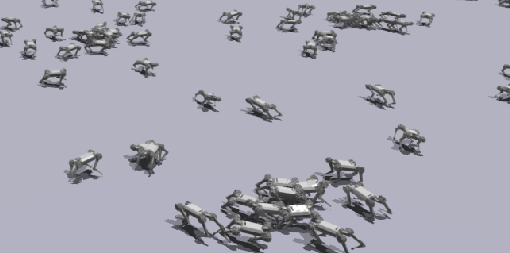}\hspace{1pt}
    \includegraphics[width=0.2434\textwidth]{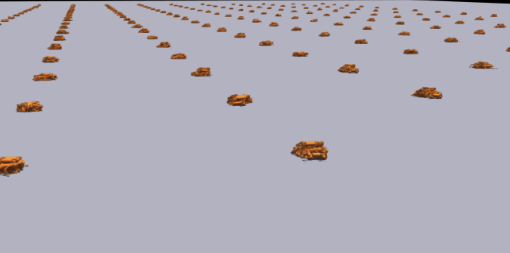}\hspace{1pt}
    \vspace{2pt}

    \includegraphics[width=0.2434\textwidth]{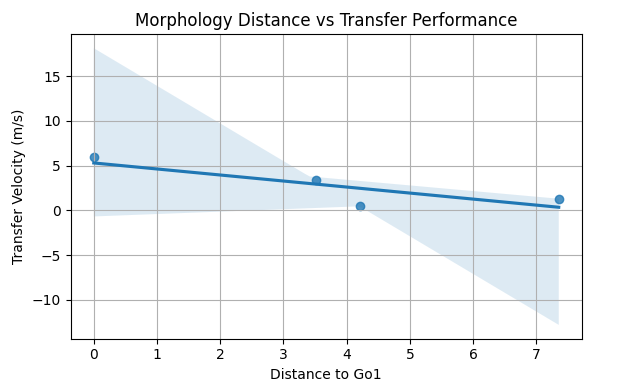}\hspace{1pt}
    \includegraphics[width=0.2434\textwidth]{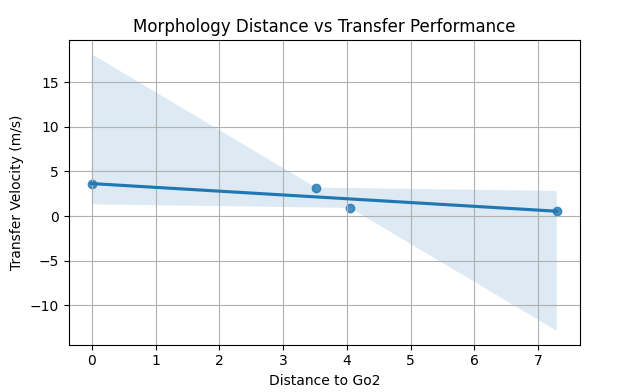}\hspace{1pt}
    \includegraphics[width=0.2434\textwidth]{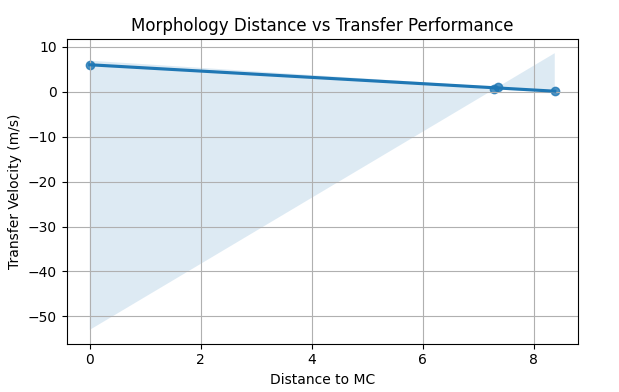}\hspace{1pt}
    \includegraphics[width=0.2434\textwidth]{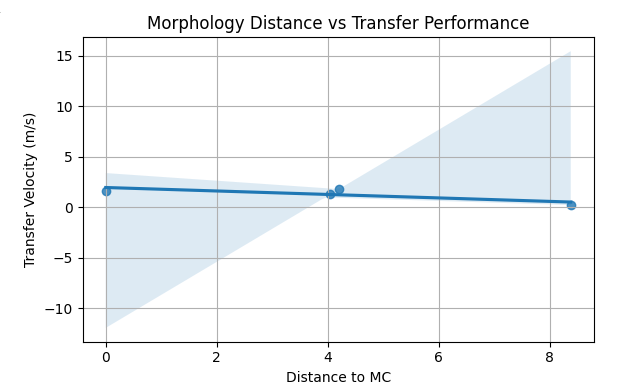}\hspace{1pt}

    \caption{Parallelized training for go1, go2, mini cheetah and A1 robot (total 4000 environments) in Isaac Gym simulator (Top row). Transfer loss and the distance correlation for \textbf{McARL}, as the morphology between the trained morphology and the unseen morphology increases, so does the transfer losses. }
    \label{fig:two_row_eight_images2}
\end{figure}

\subsection{Morphology Distance analysis and performance}
Even though McARL achieves good transfer for go2, but the performance and transfer rate decreases, even though the policies are conditioned on randomized morphology vectors but it still has some gradual reduction in transfer efficiency.

\begin{equation}
D_{\text{weighted}}(i,j) = \sqrt{ \sum_{k=1}^{14} w_k \left( \frac{z_k(i) - \mu_k}{\sigma_k} - \frac{z_k(j) - \mu_k}{\sigma_k} \right)^2 }
\label{eq:normalized distance}
\end{equation}

We define normalized and weighted euclidean distance to explain this, as the distance between the trained robot and transfer robot increase, so does the transfer loss and unstable behavior.

\begin{figure}[t]
    \centering
    \includegraphics[width=0.24\textwidth]{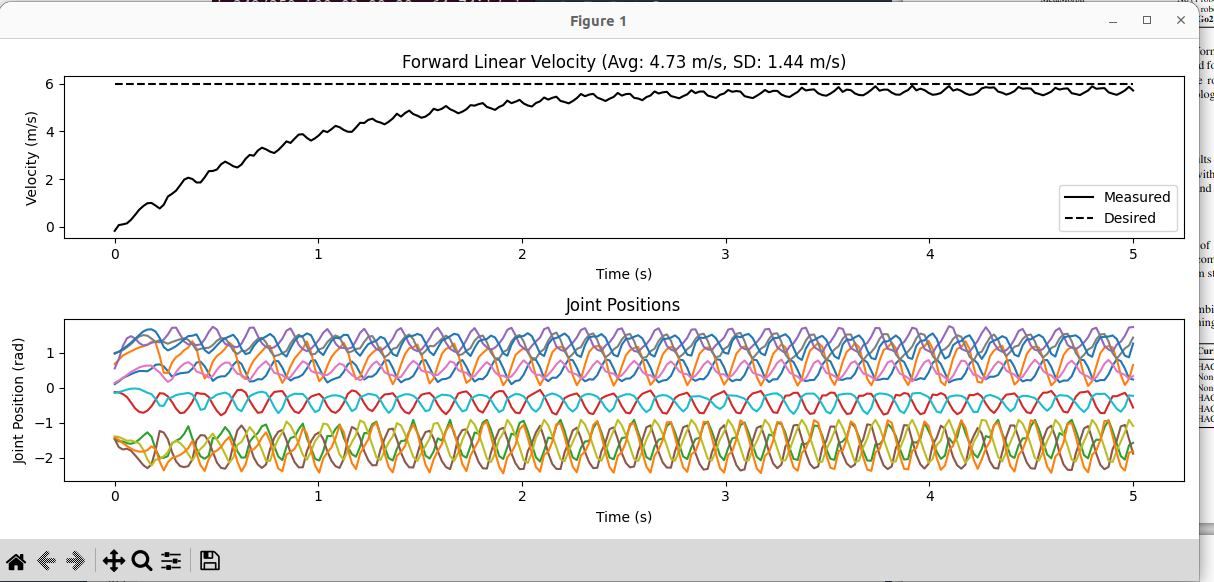}\hspace{1pt}
    \includegraphics[width=0.24\textwidth]{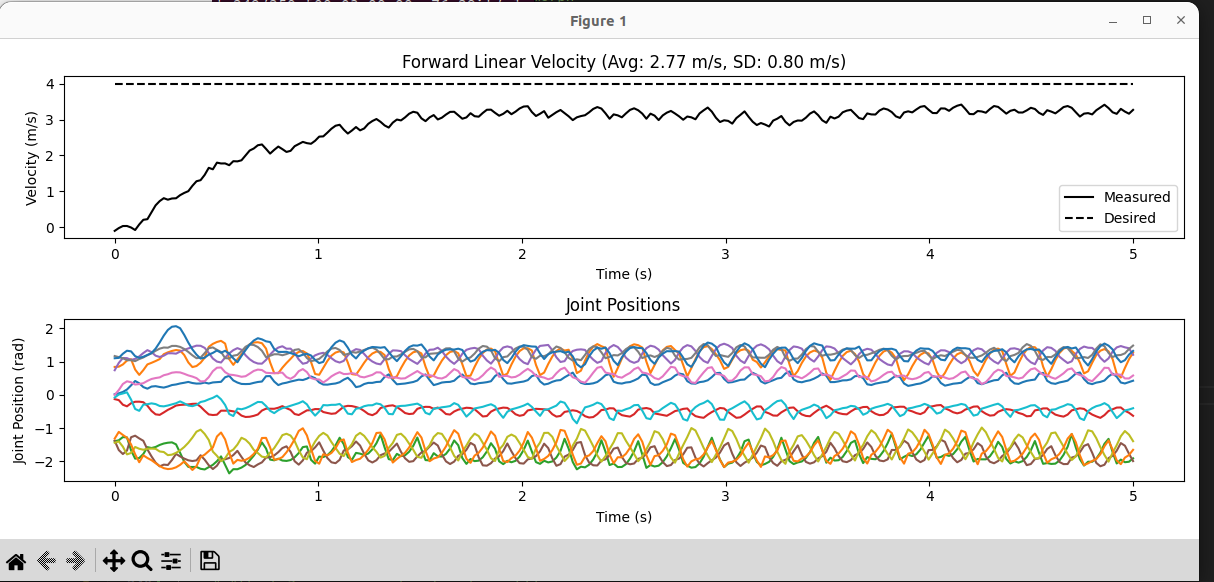}\hspace{1pt}
    \includegraphics[width=0.24\textwidth]{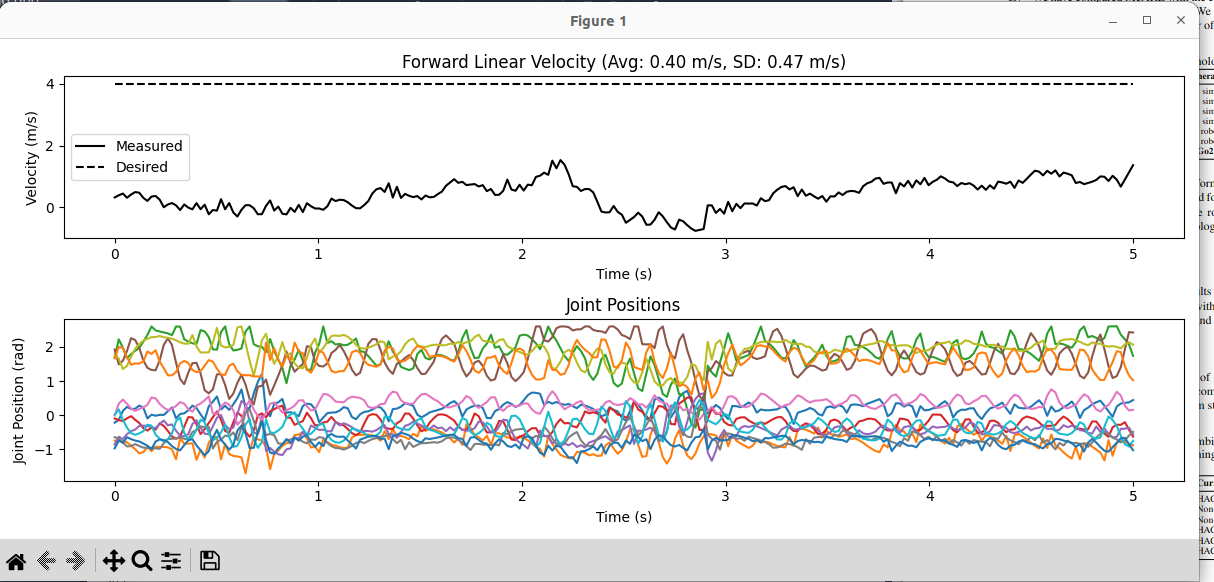}\hspace{1pt}
    \includegraphics[width=0.24\textwidth]{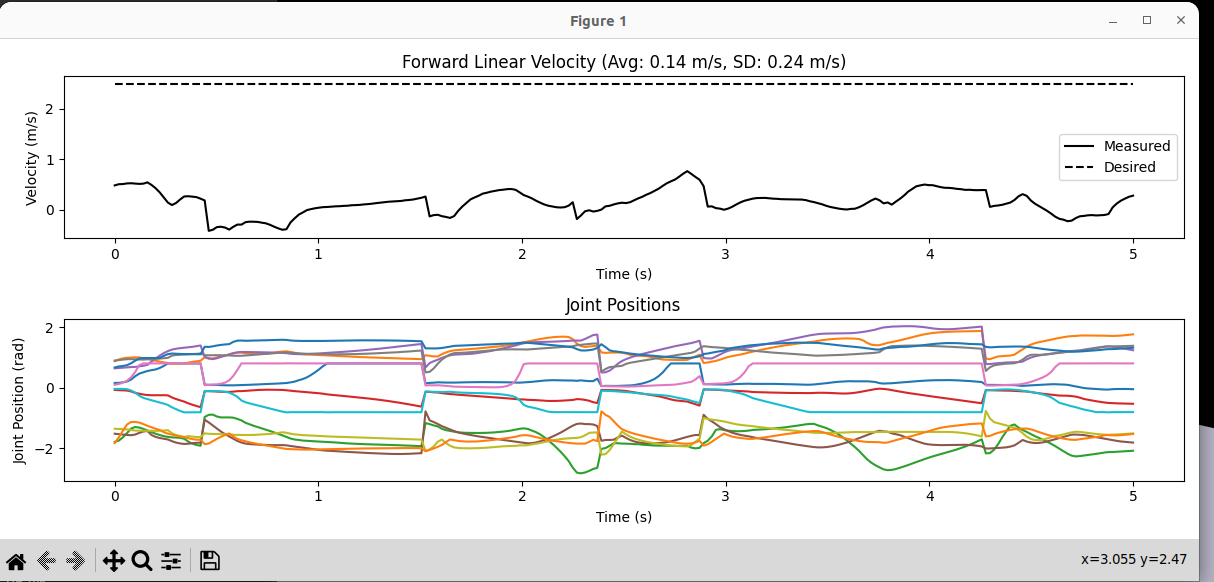}

    \vspace{2pt}
  
    \caption{PPO combination of P3 {PPO + Morphology (without control params) + HACL} performs overall best compared to other combinations like P0, P1, P2, P4, P5 and has pretty decent transfer over all other morphologies like Go1, Go2, A1 and mini cheetah robots. (Please refer to appendix for detailed analysis)}
    \label{fig:two_row_eight_images3}
\end{figure}




\subsection{Real-world Deployment}
And the demonstrate the effectiveness of our method, we did a real-world testing of our McARL controller on Go1 robot. Given fragility of the hardware and its velocity limitations for higher command velocities, so we kept our testing for 1.0-1.5 m/s command velocities. There are challenges in sim-to-real transfer like unstable gait, forward falling and very high torque etc, so in order to overcome those we fine-tuned the model to improved the robustness and stability of our model.


\section{Conclusion \& Future Work}
\label{sec:conclusion}

In this work we presented novel morphology and control based based policy learning (McARL), which highlights the importance of the ideas of morphology latent and controls parameter by integrating the them into the policy and value functions. Demonstrated the utility of McARL through zero-shot transfer on Go2, Mini Cheetah and A1 robots without the need of training and finetuning. Also tested our approach on the real world robot. Future directions that we would like to explore and expand our work McARL are i) Exploring the methods and ways to make it more robust for heavier legged morphologies as the morphology distance increases ii) Why just the control randomization leads to a conservative output iii) We would also like to expand the utility of McARL in terms of bipedal locomotion and possibly even for manipulators, iv) And incorporate McARL with the multi-modal learning.

\section{Limitations}
\label{sec:limitations}


Our McARL's performance drops slowly as the morphology distance increases from the trained robot. For example if the robot used for training is Go1, then the transfer rate increases gradually as the distance between the morphologies increases,as evident by the heat-maps (Please refer to appendix for details).There are challenged in sim-to-real transfer, McARL needs some finetuning to work robustly on the real robot (Unitree Go1 for this case). McARL demonstrates promising sim-to-sim transfer, sim-to-real deployment is challenging because of multiple reasons like sensor noise, actuator latency, and unmodeled dynamics—particularly, which can be overcome by some finetuning, when running on the Unitree Go1 robot. Our approach is specifically designed for the quadrupedal and scaling it to bipedals or manipulators would need further detailed investigation into morphology latent, reward design, and its incorporation in the actor-critic functions, which we will explore in the future work.

\clearpage



\nocite{*}

\bibliography{example}  

\clearpage
\appendix

\end{document}